\documentclass{applemlr}
\usepackage{amsmath}
\usepackage{enumerate} 
\usepackage{algorithm}
\usepackage{algpseudocode}
\usepackage{amsfonts}
\usepackage{amsthm}
\usepackage{newtxtt}
\usepackage{cleveref}
\usepackage{diagbox} 
\usepackage{colortbl}
\usepackage{amssymb}
\usepackage{xspace}
\usepackage{wrapfig}
\usepackage{adjustbox}
\usepackage{tabularx}
\usepackage{booktabs}
\usepackage{mathtools}
\usepackage{tikz}
\usepackage{enumitem}
\usepackage{silence}
\usepackage{dsfont}
\usepackage[table]{xcolor}
\usepackage[dvipsnames]{xcolor}
\usepackage{multirow}
\usepackage{makecell}
\usepackage{xfakebold}
\usepackage{amsmath,amsfonts,bm}

\def\eqref#1{equation~\ref{#1}}
\def\1{\bm{1}}

\DeclareMathAlphabet{\mathsfit}{\encodingdefault}{\sfdefault}{m}{sl}
\SetMathAlphabet{\mathsfit}{bold}{\encodingdefault}{\sfdefault}{bx}{n}

\usepackage{color}
\usepackage[dvipsnames]{xcolor}
\definecolor{JalapenoRed}{RGB}{183,21,64}
\definecolor{Belize}{RGB}{41,128,185}
\definecolor{Amour}{RGB}{238,82,83}
\definecolor{lightblue}{HTML}{dfebf7}
\definecolor{Gray}{gray}{0.93}
\definecolor{magnolia}{rgb}{0.97, 0.96, 1.0}
\definecolor{mayablue}{rgb}{0.45, 0.76, 0.98}
\definecolor{lavenderblue}{rgb}{0.8, 0.8, 1.0}
\definecolor{lavender}{rgb}{0.9, 0.9, 0.98}
\definecolor{islamicgreen}{rgb}{0.0, 0.56, 0.0}

\usepackage{stfloats}

\definecolor{textgray}{HTML}{6E6E73}
\usetikzlibrary{positioning, calc}
\usetikzlibrary{decorations.pathmorphing}

\makeatletter
\patchcmd{\wrong@fontshape}{\@gobbletwo}{}{}{}
\makeatother
\makeatletter
\AtBeginDocument{%
  \urlstyle{sf}
  
}
\makeatother

\numberwithin{equation}{section} 
\tcbuselibrary{minted}
\setminted[python]{
  linenos,
  breaklines,
  fontsize=\footnotesize,
  xleftmargin=2em
}

\definecolor{light}{RGB}{125, 125, 125}
\crefname{tcb@cnt@pbox}{code}{code}
\Crefname{tcb@cnt@pbox}{Code}{Code}
\crefname{assumption}{assumption}{assumption}
\Crefname{assumption}{Assumption}{Assumptions}

\newtcolorbox[auto counter]{pbox}[2][]{
  colback=white,
  title=Code~\thetcbcounter: #2,
  #1,fonttitle=\sffamily,
  fontupper=\sffamily,
  arc=2pt,
  colframe=bgcolor,
  coltitle=fgcolor,
  colbacktitle=bgcolor,
  toptitle=0.25cm,
  bottomtitle=0.125cm
}

\makeatletter
\newcommand\applefootnote[1]{%
  \begingroup
  \renewcommand\thefootnote{}%
  \renewcommand\@makefntext[1]{\noindent##1}%
  \footnote{#1}%
  \addtocounter{footnote}{-1}%
  \endgroup
}
\makeatother

\definecolor{cverbbg}{gray}{0.90}

\newcommand{\benchname}{\textsc{SO-Bench}}

\title{{SO-Bench}: A Structural Output Evaluation of Multimodal LLMs}

\author{
\parbox{\textwidth}{
  Di Feng$^{\circ \dagger}$, Kaixin Ma$^{\circ}$, Feng Nan$^{\circ}$, Haofeng Chen$^{\ast}$, Bohan Zhai$^{\ast}$, David Griffiths, Mingfei Gao, \\Zhe Gan$^{\ast}$, Eshan Verma$^{\ast}$, Yinfei Yang, Zhifeng Chen, Afshin Dehghan
}}

\affiliation{Apple}

\contribution{$^\circ$First authors, $^\ast$Core authors, $^{\dagger}$Project lead}

\abstract{
Multimodal large language models (MLLMs) are increasingly deployed in real-world, agentic settings where outputs must not only be correct, but also conform to predefined data schemas. Despite recent progress in structured generation in textual domain, there is still no benchmark that systematically evaluates schema-grounded information extraction and reasoning over visual inputs. In this work, we conduct a comprehensive study of visual structural output capabilities for MLLMs with our carefully designed \benchname{} benchmark. Covering four visual domains, including UI screens, natural images, documents, and charts, \benchname{} is built from over $6.5$K diverse JSON schemas and $1.8$K curated image–schema pairs with human-verified quality. Benchmarking experiments on open-sourced and frontier proprietary models reveal persistent gaps in predicting accurate, schema compliant outputs, highlighting the need for better multimodal structured reasoning. Beyond benchmarking, we further conduct training experiments to largely improve the model's structured output capability. We make the benchmark and evaluation publicly available at \url{https://github.com/apple/ml-sobench}.
}

\date{\sffamily\today}

\begin{document}

\maketitle

\section{Introduction}
\label{sec:intro}

\begin{wrapfigure}{r}{0.62\textwidth}
\centering
\vspace{-8mm}
    \includegraphics[width=1.0\linewidth]{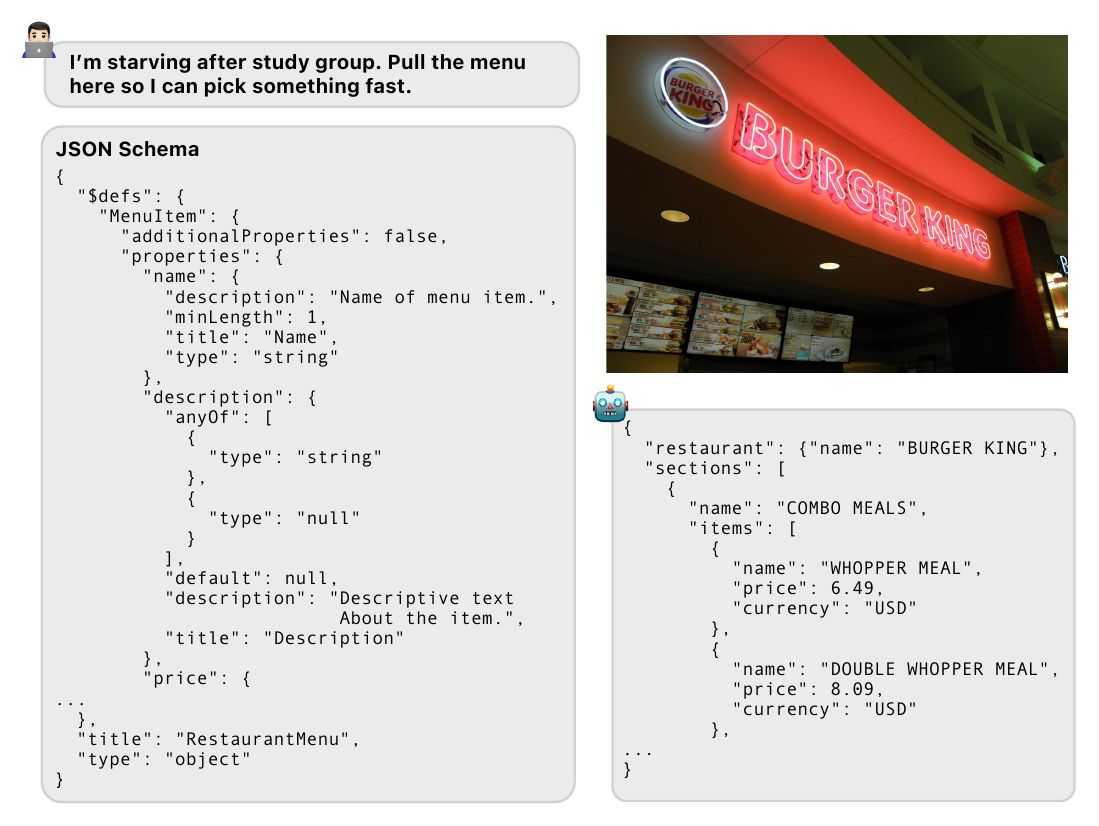}
    \vspace{-8mm}
    \caption{An example of visual structured output task. Given a customized JSON schema often specified by the downstream applications, e.g. \texttt{MenuReader}, a model is tasked to extract information from input image, following the schema definition and user instruction.}\label{fig:teaser_plot}
    \vspace{-3mm}
\end{wrapfigure}

The rapid advancement in large language models (LLMs) and multimodal large language models (MLLMs) ~\citep{mckinzie2024mm1, zhangmm1,tong2024cambrian,lillava,team2025gemma,bai2025qwen2,team2025kimi,wang2025internvl3} has expanded their role beyond natural language or image understanding and generation. Increasingly, these models are leveraged in agentic applications, such as web automation~\citep{yang2025ferret,wang2025ui,zhanghengferret}, data extraction~\citep{mixtral_ocr,niu2025mineru2,cui2025paddleocr}, and tool use~\citep{wang2024gta,chen2023fireact,lu2025toolsandbox}, where outputs are consumed not by humans but by downstream systems, controllers, or APIs. In this context, a model must produce outputs that are not only semantically correct, but also structurally valid, conforming exactly to a predefined customized input schema, typically in JSON formats. This \textit{structured outputs} formulation enables LLMs or MLLMs to interface reliably with external systems, perform programmatic reasoning, and execute grounded actions. Due to its importance, several industrial API extensions have incorporated structured output modes (e.g., OpenAI~\citep{openai_api}, Google Gemini~\citep{gemini_api}, Anthropic~\citep{claude_api}.)

While structured output from text-only inputs~\citep{yang2025structeval, tang2023struc, gu2024structext, geng2025jsonschemabench} has received growing attention, the multimodal domain, especially the visual structured output capability of MLLMs, remains largely underexplored. Given visual inputs (e.g., natural images, screens, charts) and customized schemas (often specified by downstream applications), visual structured output involves extracting information from images, and aligning them with hierarchical schemas and user instructions. For instance, adding the product labels shown on a UI screen to the shopping cart, or saving the information from the menu to the notes apps, illustrated in Figure~\ref{fig:teaser_plot}. Prior visual-language works such as Pix2Struct~\citep{lee2023pix2struct}, Image2Struct~\citep{roberts2024image2struct}, and IR3D-Bench~\citep{liu2025ir3d} have made progress toward structured visual understanding, but they typically emphasize text captioning or semantic parsing with limited domains rather than rigorous customized schema-level adherence with diverse real-world scenarios. Most relevant to visual structured output is key information extraction (KIE) in document parsing and OCR understanding~\citep{yang2024cc,ouyang2025omnidocbench}, where a MLLM is tasked to extract information for pre-defined keywords. However, KIE focuses on semantic parsing with simple input fields, ignoring the diversity of real-world schemas with nested, complex structure. As a result, there is no study that systematically quantifies how well MLLMs can produce structured, schema-compliant outputs grounded in visual evidence.

To address these limitations, we conduct a comprehensive study of visual structured output capability for MLLMs. To this end, we build \benchname{}, a visual structured output benchmark with nearly $1.8$K diverse, high-quality samples, each image paired with a unique JSON schema and user instruction, sourced from a pool of over $112$K diverse images in four domains (natural images, UI screens, documents, and charts), $6.5$K real-world and synthetic schemas, and $60$K user profiles. When constructing the benchmark dataset, we carefully design a multi-stage auto-labeling pipeline, including (1) accurate image-schema association through multimodal embedding search, (2) synthetic schema generation with multi-image grouping, and (3) human-in-the-loop progressive response generation and refinement with a critic model. In addition, human domain experts conduct inspection at each stage to ensure data quality. To quantify a model’s structured output capability, we develop an evaluation pipeline that measures both a model’s ability to follow schema instructions and its behavior under constrained decoding. The evaluation decomposes performance into schema adherence, structural fidelity, and value-level accuracy (with exact match or fuzzy match), enabling fair comparisons of different schema responses.

Based on \benchname{}, we conduct comprehensive benchmarking experiments across a wide range of MLLMs of different scales, including open-sourced and proprietary models. Our results reveal that small models (e.g., with 3B and 7B scales) exhibit substantial gaps in structured generation and schema compliance compared to their proprietary counterparts. Though the strongest frontier models (e.g., GPT-5 and Gemini 2.5-Pro) show good performance in schema following with over $95\%$ accuracy, there is a large room of improvements for predicting fully correct structured outputs, with only up to nearly $19\%$ accuracy (fuzzy match). Ablation studies further illustrate that a model’s visual structured output ability highly correlates with its competence in visual instruction following, agentic tool invocation and parameter filling, and general visual knowledge.

To study how training impacts models' visual structured output capabilities, we further construct a large-scale training set at the post-training stage using the same data labeling pipeline, and train MLLMs with supervised finetuning (SFT) and Reinforcement Learning with Verifiable Rewards (RLVR). Experimental results show that both SFT and RLVR enhances schema validation and field matching accuracy up to $20\%$ and $13\%$, respectively, underscoring the importance of targeted supervision for improving structured reasoning.

\noindent In summary, our contributions are three-folds:

\begin{itemize}
  \item A diverse, high-quality benchmark, \benchname{}, to quantify structured output capabilities of MLLMs;
  \item Comprehensive experiments to show the limitation of existing MLLMs in visual structured outputs;
  \item Model training experiments to largely enhance the structured output capabilities.
\end{itemize}
The benchmark and evaluation pipeline are publicly available at \url{https://github.com/apple/ml-sobench}.
\section{Related Works}
\label{sec:related_works}
\textbf{Structured Output Benchmarks.}
Several benchmarks have been recently proposed to quantify a LLM's capability to generate structured output with \textit{text-only} inputs. For example, StructEval~\cite{yang2025structeval} benchmarks text-to-structure generation across formats like JSON, YAML, and HTML, emphasizing format fidelity. JSONSchemaBench~\cite{geng2025jsonschemabench} extends this by testing constrained decoding on thousands of real-world JSON schemas, measuring coverage and quality across diverse constrained decoding frameworks. StructBench~\cite{tang2023struc} auto-generates structure-rich text tasks such as table-to-JSON or LaTeX conversion. In contrast to text-only applications, visual structured output focuses on extracting image information following the pre-defined schemas. Pix2Struct~\cite{lee2023pix2struct} pretrains an image-to-text model that parses screenshots to HTML, improving layout understanding. Image2Struct~\cite{roberts2024image2struct} targets structured extraction from rendered data (webpages, formulas, music scores), yet only focuses on digital images with limited domains. IR3D-Bench~\cite{liu2025ir3d} tests structured scene reconstruction from pure synthetic 3D renderings. Overall, existing studies either omit image inputs or cover narrow visual domains, leaving the broader capability of  MLLMs in generating diverse, schema-grounded outputs under-explored.

\paragraph{Agentic Tool Use Benchmarks.}
\benchname{} also relates to agentic tool-use evaluations that quantify structured reasoning and function-call accuracy. The Berkeley Function Calling Leaderboard (BFCL)~\cite{patilberkeley} and Tau-Bench~\cite{yao2024tau} assess agents’ ability to produce executable structured calls and manage multi-step tool interactions, while ToolVQA~\cite{yin2025toolvqa} extends this concept to visual question answering with external tools. Unlike these benchmarks, which focus on textual tool APIs or with limited number of tools (e.g., $14$ tools in GTA~\citep{wang2024gta}), \benchname{} evaluates multimodal structured generation grounded in visual inputs with large-scale diverse JSON schemas, bridging perception and structured reasoning.

\paragraph{OCR Understanding.}
Visual structured output connects to document understanding and OCR-rich tasks, as it requires a model to correctly parse textual information from images. Benchmarks such as OCRBenchV2~\citep{fu2024ocrbench}, CC-OCR~\citep{yang2024cc}, and OmniDocBench~\citep{ouyang2025omnidocbench} focus on extracting text and layout from scanned or synthetic documents, often formulating key information extraction (KIE) tasks with simple predefined templates in one-layer dictionary. In contrast, \benchname{} focuses on distinct, realistic JSON schemas with nested structures, aligned with application-driven agentic tool-use requirements. 

\section{The \benchname{} Dataset}
\label{sec:benchmark}
\subsection{Overview}~\label{sec:benchmark_overview}
We formulate the visual structured output problem as follows: given an input image $I$, a pre-defined JSON schema $S$, and a user instruction $X$, 
a model is expected to generate structured output $Y$ that both (1) conforms syntactically to $S$ and (2) semantically reflects the information extracted from $I$ and $X$. 
For a multimodal LLM, the structured output is generated auto-regressively with the predicted probability $p(Y|I,X,S)$. A JSON schema is a nested dictionary specifying keys, 
data types, and object hierarchy following the standard~\citep{jsonschema}. A user instruction provides guidance on the information to be extracted from the input images, which can be precise and descriptive, or concise and ambiguous, 
for example, ``Help me save \textit{this} poster''.

The benchmark should be diverse in order to test the general visual structured output capabilities of MLLMs. The diversity originates from images (domain coverage and visual representations) and JSON schemas (e.g., schema complexity, field type variety), which will be introduced in Section~\ref{sec:data_collection}.
The main challenge for building such a benchmark is how to associate an image with the representative JSON schema, and generate its structured output response accurately and efficiently. To tackle this challenge, we develop a multi-stage auto-labeling pipeline 
with human in-the-loop detailed in Section~\ref{sec:data_curation}.

\begin{figure*}[t!]
    \centering
    \includegraphics[width=1.00\linewidth]{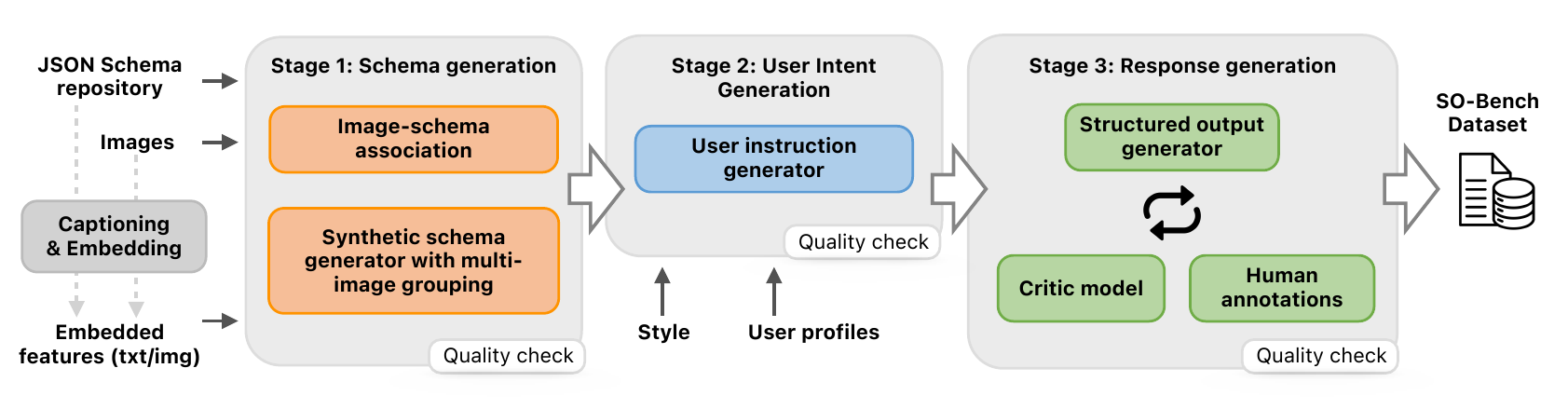}
    \vspace{-3mm}
    \caption{Overview of the multi-stage data generation pipeline for \benchname{}, including schema generation, user intent generation, and response generation stages. At each stage, we leverage proprietary frontier models (e.g., GPT-5 and Gemini-2.5-Pro) as generators with careful prompt design. Data from each stage is checked with human domain experts before passing to the next stage. Before the schema generation stage, input images and JSON schemas are embedded through a CLIP model for embedding search. Details of the pipeline is introduced in Section~\ref{sec:data_curation}.}
    \label{fig:data_generation}
\end{figure*}

\subsection{Data Collection}~\label{sec:data_collection} 
To ensure broad visual coverage, \benchname{} aggregates image data from ten recent public datasets spanning four domains with different image resolutions and aspect ratios: (1) UI (mobile, web, and desktop screens), (2) natural images, (3) documents, and (4) charts. 
These domains jointly cover a wide range of real-world structured extraction scenarios, such as menu parsing, receipt understanding, poster collection, UI layout analysis, document parsing, and chart interpretation. 
We prioritize English-language data and selectively down-sample test sets to maximize visual and semantic diversity. Figure~\ref{fig:dataset_composition} summarizes the data composition and domain distribution. Regarding JSON schema collection, we curated a large and diverse schema repository covering both real-world and synthetic cases. The repository includes a subset of the real-world JSON schemas from JsonSchemaBench~\citep{geng2025jsonschemabench}, which spans domains such as function signatures, service APIs, and system configurations. Out of $11$K full schema repository, we keep $6$K unique schemas related to visual applications. We additionally generate $500$ synthetic JSON schemas by prompting a proprietary model for specific use cases (e.g., reading receipt, parsing nutrition labels). This collection serves as a JSON schema databank from which image–schema pairs are constructed.

\paragraph{Embedding.}
To enable efficient schema–image retrieval and matching (detailed in Section~\ref{sec:data_curation}), we embed both images and JSON schemas using an off-the-shelf CLIP model~\citep{radford2021learning}.  
For an Image $I$, we first use a proprietary model (e.g., GPT-5 or Gemini-2.5-Pro) to generate its caption $T$, and then employ the CLIP's image and text encoders to extract visual and textual embeddings, denoted by $E_I$ and $E_T$, respectively.
As for a JSON schema $S$, we prompt with another proprietary model to summarize its fields, purpose, and example use cases with a canonical name (e.g., \texttt{ReceiptExtractor}, \texttt{MenuReader}). These textual descriptions are embedded with the CLIP text encoder, denoted by $E_S$.

\subsection{Data Curation}~\label{sec:data_curation}
Figure~\ref{fig:data_generation} illustrates the the full data curation workflow, which can be divided into three stages. In the schema generation stage, a JSON schema is linked to the input image, 
either selected from the schema repository (Section~\ref{sec:image_schema_association}) or generated on-the-fly with multi-image grouping (Section~\ref{sec:multi_image_schema_gen}). In the second stage, 
user instructions are added to the image-schema pairs to simulate realistic human-agent interaction patterns. Inspired by the persona-based prompt generation~\citep{ge2024scaling}, we first synthesize $60$K 
diverse user profiles varying in age, occupation, and locale. For each image–schema pair, a user profile and the chat style are sampled randomly as context information for the user intent, leading to a rich spectrum of instruction styles, including conversational, 
direct, ambiguous, or even dialectal to reflect natural user diversity. At the response generation stage, the structured output is generated and iteratively refined, based on a critic model and human annotators' feedback. 
To ensure data quality, a group of eight human domain experts check and filter the intermediate outputs, before proceeding to the next stages.

\subsubsection{Image-Schema Association}~\label{sec:image_schema_association}
Given an image embedding, we retrieve the top-$k$ most relevant JSON schemas from the schema databank based on the visual-textual multimodal nearest neighbor search. Given an image $I$ and schema $S$, its similarity score is computed as the weighted sum of cosine similarities between image/caption embeddings and schema embeddings, written as: 
\begin{equation}
    \text{sim}(I, S) = w_1 \cdot \text{cos}(E_I, E_S) + w_2 \cdot \text{cos}(E_T, E_S).
\end{equation}
From these $k$ candidates, a large multimodal model is prompted to select the best-matching schema, ensuring semantic alignment and structural appropriateness. Typically, we set $k=20$, and interleave schema selection from the model or random selection to increase data diversity.

Note that the semantic search approach described above could still end up with less-relevant image-schema associations. We performed human quality check and filter out those samples before proceeding to the next data generation stage.

\subsubsection{Schema Generation with Multi-image Grouping}~\label{sec:multi_image_schema_gen}
To further enhance schema diversity and structural depth, we introduce a multi-image schema generation pipeline. For each query image, we identify its top-$m$ nearest neighbor images in the embedding space (based on both image and caption similarity). 
The selected image cluster (in our experiment $m=3$) is jointly passed to a schema generator that proposes a unified nested schema capturing the shared structure across the images, e.g., multi-item menus or multi-section forms. 
This approach effectively generalizes synthetic schema generation beyond single-image templates. Formerly, given images $i$ and $j$, their similarity score is computed as: 
\begin{equation}
    \text{sim}(I_i, I_j) = w_1 \cdot \text{cos}(E_{I_i}, E_{I_j}) + w_2 \cdot \text{cos}(E_{I_i}, E_{T_j}) + w_3 \cdot \text{cos}(E_{T_i}, E_{I_j}) + w_4 \cdot \text{cos}(E_{T_i}, E_{T_j}).
\end{equation}

\subsubsection{Progressive Response Generation and Refinement}~\label{sec:response_gen}
For each triplet of image, schema, and user intent, we generate the structured output using our response generation pipeline. When available, auxiliary signals such as OCR texts, layout metadata, or HTML structures (for UI images) provided by the original benchmark datasets are used as guidance to improve fidelity. 
The generated JSON outputs are validated against schema constraints to ensure syntactic correctness and logical completeness. Besides, we introduce a hybrid automated and manual review loop. A critic–refiner workflow, powered by LLM-based validators, examines each structured output 
for schema validity and semantic consistency and summarizes suggestions of improvements. Invalid or sub-optimal outputs are regenerated up to three times considering the feedback from the critic model. During this process, human experts conduct manual inspections and provide feedback
for the output refinement.

\begin{figure*}[t]
    \centering
    \begin{subfigure}[t]{0.42\textwidth}
        \centering
        \includegraphics[width=\linewidth]{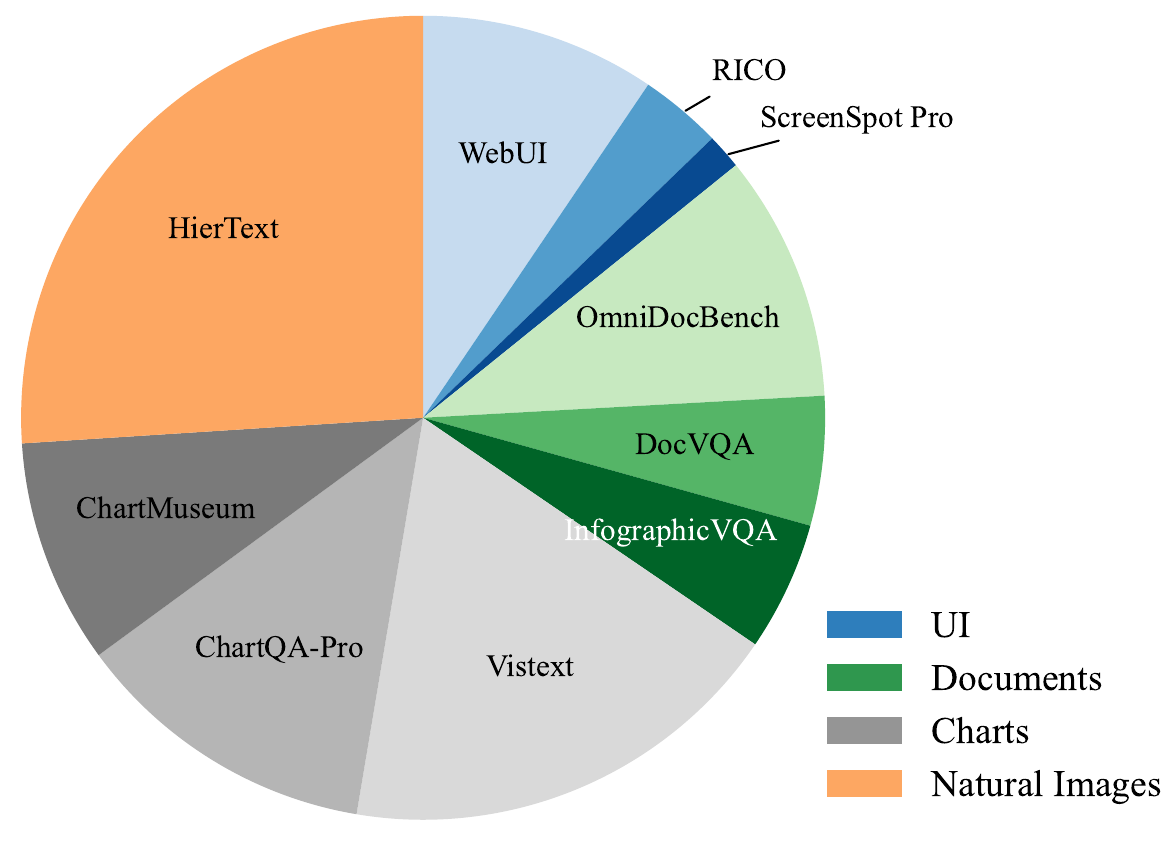}
        \caption{Image category composition.}
        \label{fig:dataset_composition}
    \end{subfigure}
    \begin{subfigure}[t]{0.36\textwidth}
        \centering
        \includegraphics[width=\linewidth]{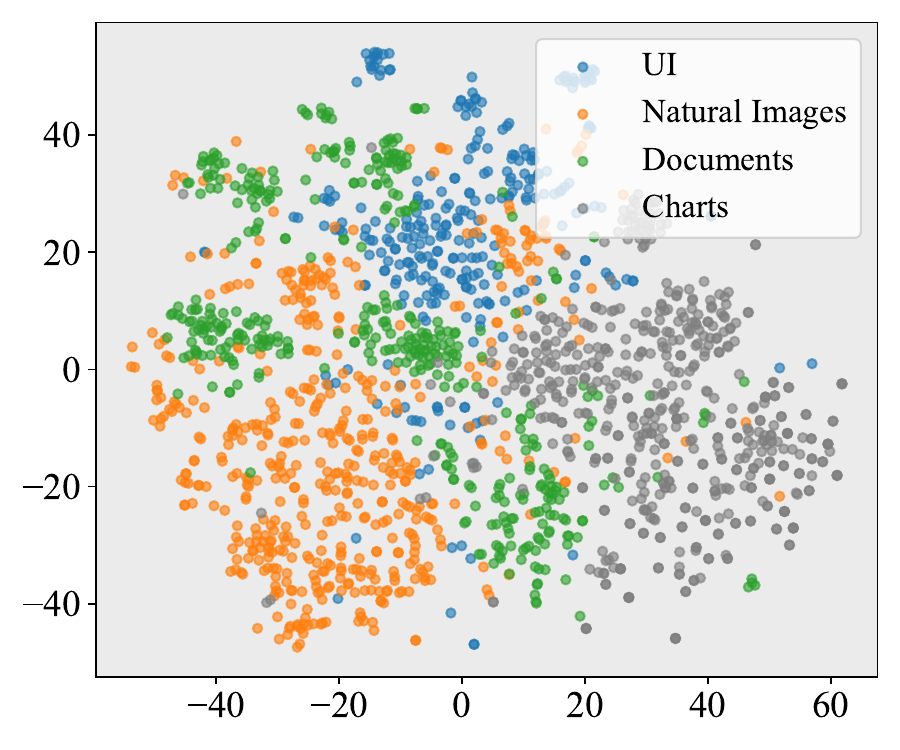}
        \caption{t-SNE Plot for image features.}
        \label{fig:image_tsne}
    \end{subfigure}
    \caption{\textbf{(a).} \benchname{} is collected from four categories: (1) UI: RICO~\citep{deka2017rico}, WebUI~\citep{wu2023webui}, ScreenSpot Pro~\citep{li2025screenspot}); (2) Documents: OmniDocBench~\citep{ouyang2025omnidocbench},
    DocVQA~\citep{mathew2021docvqa}, InfographicVQA~\citep{mathew2022infographicvqa}; (3) Charts: ChartQA-Pro~\citep{masry2025chartqapro}, ChartMuseum~\citep{tang2025chartmuseum}; and (4) Natural images: HierText~\citep{long2022towards}. \textbf{(b).} Image feature distribution. We use CLIP~\citep{radford2021learning} image embedder to embed all images, and show their t-SNE features. The distribution from each image category is different, indicating the visual diversity.}
    \label{fig:image_feature_distribution}
\end{figure*}

\subsection{Dataset Statistics}
\begin{figure*}[t]
    \centering
    \begin{subfigure}[t]{0.32\textwidth}
        \centering
        \includegraphics[width=\linewidth]{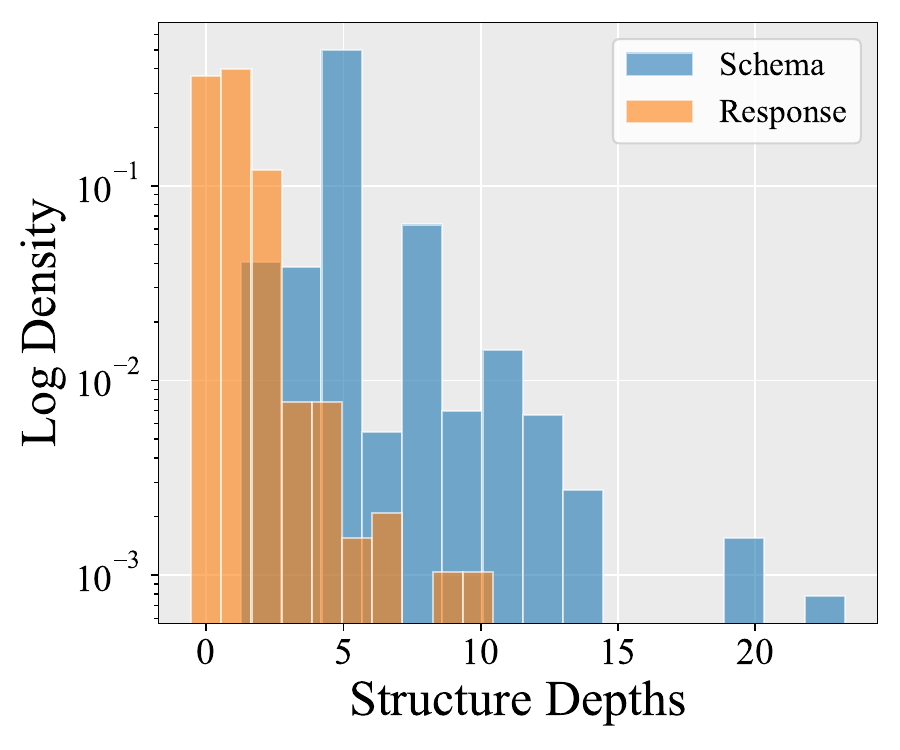}
        \caption{Structure depth distribution.}
        \label{fig:depth_distribution}
    \end{subfigure}
    \hfill
    \begin{subfigure}[t]{0.32\textwidth}
        \centering
        \includegraphics[width=\linewidth]{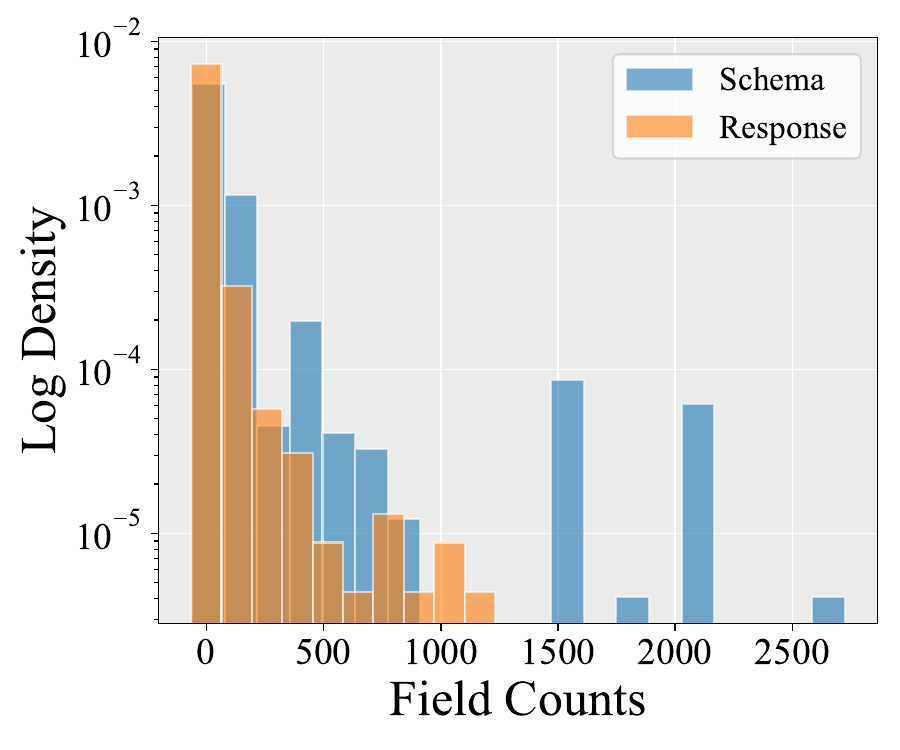}
        \caption{Field count distribution.}
        \label{fig:field_count_distribution}
    \end{subfigure}
    \hfill
    \begin{subfigure}[t]{0.32\textwidth}
        \centering
        \includegraphics[width=\linewidth]{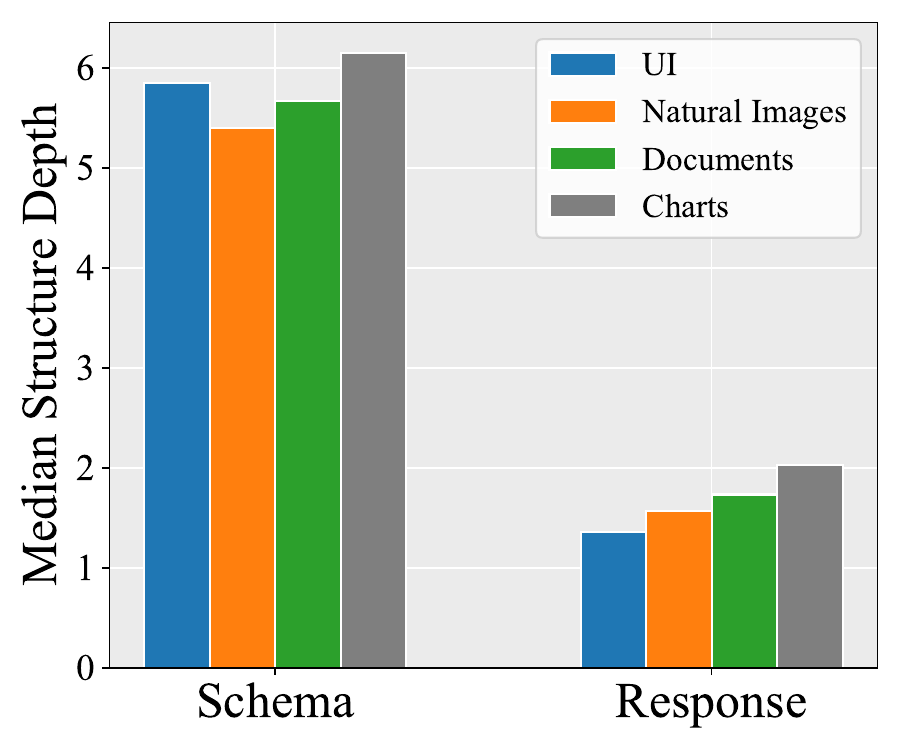}
        \caption{Median Structure depth.}
        \label{fig:depth_comparison}
    \end{subfigure}
    \caption{JSON schema data statistics in \benchname{}. The input JSON schemas are more complex than output formats in terms of the number of nested structure depths and the number of fields. Regarding image category, chart data shows the most complex schema structure.}
    \label{fig:data_statistics}
\end{figure*}

\begin{figure}[t!]
    \centering
    \includegraphics[width=1.0\linewidth]{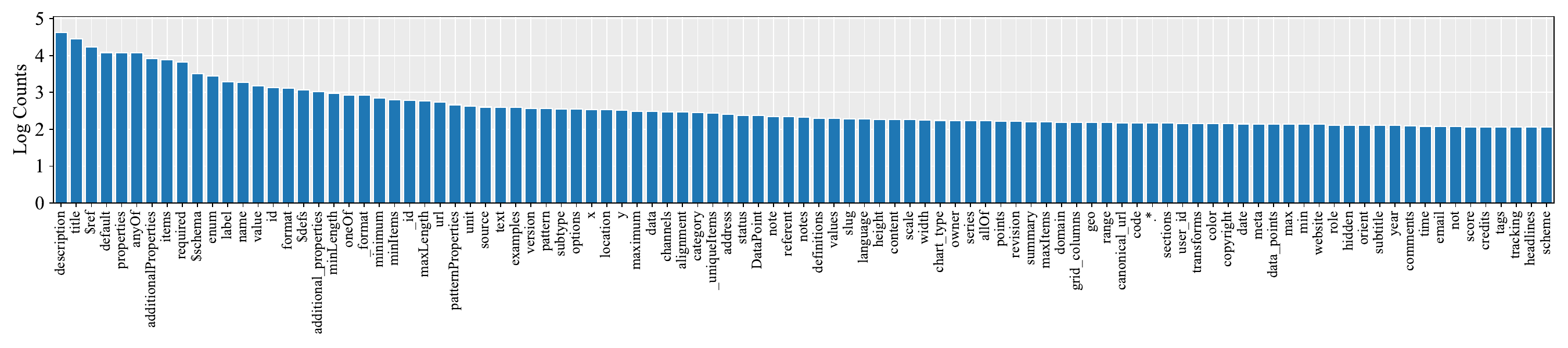}
    \caption{Top $100$ frequent feature counts in the input JSON schemas (best view with magnificence).}
    \label{fig:feature_counts_distribution}
\end{figure}

To characterize structural complexity, we compute both the depth of nested hierarchies and the number of fields in each JSON schema and corresponding structured output. Figure~\ref{fig:depth_distribution} and Figure~\ref{fig:field_count_distribution} present the distributions of structure depth and field count, respectively. In general, input JSON schemas exhibit higher complexity than the generated structured outputs, ranging from a single-level schema with one field to as deep as $22$ levels and over $2$K fields, reflecting the rich diversity of schema structures in \benchname{}. Figure~\ref{fig:depth_comparison} further compares the median structure depth across four image domains, showing that charts tend to involve slightly deeper hierarchies than other categories. Figure~\ref{fig:image_tsne} visualizes the t-SNE projection of image embeddings obtained from the CLIP image encoder. Images from different domains form distinct clusters, confirming the visual diversity of the dataset. Finally, we plot the top $100$ most frequent feature names appearing in the JSON schema repository in Figure~\ref{fig:feature_counts_distribution}.

\subsection{Evaluation Metrics}
\label{sec:eval}
We used abstract syntax tree (AST) evaluation method following BFCL~\citep{patilberkeley}. In particular, given the structured output from the model and ground truth in dict format, we iteratively compare all keys in the dict and their corresponding values, and recursively compare all the nested dictionaries. For all primitive types (int, float, str, list), we used \textit{exact match} by default. Since not all information in the ground truth is directly relevant to the user intent,  and some visual information can be hard to extract exactly (e.g., blurry, truncated images), we further define two other matching strategies for primitive types: 1) we allow \textit{fuzzy match} if the ground truth information does not include explicit texts shown in the image (e.g., exact data value in a line plot).  Here, we adopt normalized edit distance to compare strings and relative error to compare int and float. 2) We also label \textit{ignore} for certain fields, if the key is not a required field in the schema definition and  it is not directly related to the user query (e.g., the style of an UI screenshot when asking about content). Here, we simply assign a matching score of $1$, if the fuzzy match score is above a pre-defined threshold and skip comparisons for ignored fields. To enable such fine-grained evaluation, we also add an evaluation label generation step in our pipeline, where we provide the image, schema, user intent and ground-truth response (after final refinement), and prompt a proprietary MLLM to generate a matching type for each primitive field (e.g., one of \{exact, fuzzy, ignore\}). The pseudo code of our evaluation function is presented in Algorithm \ref{alg:field-match}.

We report the following three metrics in our experiments. (1) \textbf{Schema Validation Accuracy:} the percentage of outputs that are valid w.r.t the given schema definition. (2) \textbf{Field Matching Accuracy (FMA):} Let $\mathcal{F}(D)$ denote all fields in nested dictionary $D$ (including both intermediate nested structures and leaf fields). For ground truth $G^{(k)}$ and output $O^{(k)}$ across $N$ examples: 
\begin{equation}
    \text{FMA} = \frac{\sum_{k=1}^{N} |\{f \in \mathcal{F}(G^{(k)}) : \exists f' \in \mathcal{F}(O^{(k)}), \text{Match}(f, f')\}|}{\sum_{k=1}^{N} |\mathcal{F}(G^{(k)})|}, 
\end{equation}\label{eq:fma}
where $\text{Match}(f, f')$ returns true for exact/fuzzy matches, with nested structures matching when all their subfields match. (3) \textbf{Full Structure Matching Accuracy (FSMA):}
\begin{equation}
    \text{FSMA} = \frac{1}{N} \sum_{k=1}^{N}  1 \left[\forall f \in \mathcal{F}(G^{(k)}), \exists f' \in \mathcal{F}(O^{(k)}) : \text{Match}(f, f')\right].
\end{equation}\label{eq:fsma}

\section{Experiments and Results}
\label{sec:experiments}
In this section, we first report our benchmarking results in Section~\ref{sec:main_results} and Section~\ref{sec:ablation}, and then introduce our training experiments in Section~\ref{sec:training_studies}. Unless mentioned otherwise, we evaluate a model's visual structured output capability with a detailed system prompt illustrating the task instruction. To do that, we serialize the input JSON schema, append it with the user intent to prompt the models for response. Detailed prompt is provided in Appendix~\ref{app:instruction_prompt}.

\begin{table*}[t!]
\centering
\resizebox{1.0\linewidth}{!}{
\begin{tabular}{lccccc}
\toprule
Models  & Schema Val. & Field Match (Exact) & Full Match (Exact) & Field Match (Fuzzy) & Full Match (Fuzzy) \\ \midrule
Qwen3-VL (2B)~\citep{yang2025qwen3} & $16.28$ & $36.42$ & $2.50$ & $39.53$ & $6.36$ \\
Qwen2.5-VL (3B)~\citep{bai2025qwen2} & $60.71$ & $40.58$ & $1.12$ & $41.59$ & $2.68$ \\
Qwen3-VL (4B)~\citep{yang2025qwen3} & $57.32$ & $53.00$ & $3.76$ & $53.49$ & $6.78$ \\
Gemma-3 (4B)~\citep{team2025gemma} & $61.47$ & $30.40$ & $0.77$ & $31.10$ & $1.43$ \\
Intern3.5-VL (4B)~\citep{wang2025internvl3} & $71.91$ & $50.55$ & $1.68$ & $51.53$ & $2.96$ \\
\midrule
Phi-4-Vision (5.6B)~\citep{abouelenin2025phi} & $22.04$ & $27.27$ & $0.46$ & $27.78$ & $0.72$ \\
Qwen3-VL (8B)~\citep{yang2025qwen3} & $54.72$ & $53.24$ & $5.24$ & $55.06$ & $8.25$ \\
Pixtral-2409 (12B)~\citep{agrawal2024pixtral} & $77.16$ & $43.19$ & $1.12$ & $45.64$ & $3.30$ \\
LLama-4-Scout (17B-16E)~\citep{llama4} & $83.78$ & $52.80$ & $2.74$ & $54.16$ & $5.54$ \\
\midrule
Gemma-3 (27B)~\citep{team2025gemma} & $82.93$ & $47.06$ & $1.68$ & $47.33$ & $3.01$ \\
Qwen2.5-VL (32B)~\citep{bai2025qwen2} & $87.16$ & $57.28$ & $3.84$ & $58.84$ & $6.91$\\
Qwen3-VL (32B)~\citep{yang2025qwen3} & $47.71$ & $58.38$ & $5.74$ & $59.35$ & $9.57$ \\
Intern3.5-VL (38B)~\citep{wang2025internvl3} & $89.32$ & $57.68$ & $3.08$ & $58.01$ & $5.35$ \\
Qwen2.5-VL (72B)~\citep{bai2025qwen2} & $87.39$ & $57.86$ & $4.22$ & $58.91$ & $9.25$\\
\midrule
Claude-4.5-Haiku~\citep{claude-4-5} & $95.70$ & $61.79$ & $3.63$ & $62.44$ & $6.93$ \\
Claude-4.5-Sonnet~\citep{claude-4-5-sonnet} & $96.50$ & $62.32$ & $5.15$ & $62.67$ & $8.74$ \\
GPT-4o-mini~\citep{hurst2024gpt} & $83.64$ & $50.91$ & $2.67$ & $52.74$ & $5.82$ \\
GPT-4o~\citep{hurst2024gpt} & $76.94$ & $59.73$ & $4.96$ & $60.91$ & $10.39$ \\
GPT-5-mini~\citep{gpt5} & $\bf 98.70$ & $58.59$ & $5.09$ & $60.29$ & $10.07$ \\
GPT-5~\citep{gpt5} & $96.38$ & $61.17$ & $5.22$ & $62.74$ & $11.60$ \\
Gemini-2.5-flash~\citep{comanici2025gemini} & $91.69$ & $64.93$ & $6.74$ & $66.32$ & $11.31$ \\
Gemini-2.5-pro~\citep{comanici2025gemini} & $97.74$ & $\bf 71.46$ & $\bf 11.38$ & $\bf 73.14$ & $\bf 18.91$ \\
\bottomrule
\end{tabular}
}
\caption{A comparison of structured output performance among different models. We cluster models based on their model size (active parameters), as well as open-sourced or proprietary models.} 
\label{tab:main_results}
\end{table*}

\subsection{Main Results}~\label{sec:main_results}
Table~\ref{tab:main_results} compares the structured output performance across several state-of-the-art open-source and large proprietary models. We report three metrics, namely, schema validation accuracy, field matching accuracy (exact and fuzzy), and full structure matching accuracy (exact and fuzzy), as introduced in Section~\ref{sec:eval}. Among all models, Gemini-2.5-Pro achieves top results across all metrics, with a schema validation accuracy reaching nearly $98\%$, demonstrating strong schema adherence capabilities. GPT-5 also shows a clear improvement over its predecessor GPT-4o with over $15\%$ higher schema validation accuracy. However, all models exhibit relatively lower performance in field matching. For field match (fuzzy), only Gemini-2.5-Pro achieves accuracy higher than $70\%$, and no model achieves above $20\%$ on Full Match (fuzzy), indicating substantial room for improvement in producing fully compliant structured outputs. Upon further inspection of errors, we also found that the large gap between field match and full match accuracies are often due to the model predicting semantically correct answers for certain fields, which is beyond what fuzzy match can capture. We leave semantic matching methods out from our metrics for simplicity and reproducibility, and we think developing more flexible matching functions could be an interesting future work direction.
\subsection{Detailed Analysis}~\label{sec:ablation}
\subsubsection{Metrics Correlation Analysis}
To better understand how \benchname{} relates to existing benchmarks, we analyze its correlation with several public datasets that capture complementary aspects of multimodal reasoning. To do that, we compute Pearson correlation coefficients between the \benchname{} metrics and representative benchmark metrics among several models in Table~\ref{tab:main_results}. The results are summarized in Table~\ref{tab:correlation_metrics} (Appendix~\ref{app:ablation_studies} provides more details). Strong and statistically significant correlations with low $p$ values are highlighted in blue. From the results, \benchname{} shows the strongest alignment with BFCL~\citep{patilberkeley}, LiveBench (Coding)~\citep{white2024livebench}, MMMU~\citep{yue2024mmmu}, and MIABench~\citep{qian2024mia}, indicating that structured output performance is closely linked to a model’s agentic reasoning and tool use, general vision knowledge, and visual instruction-following abilities, respectively. Though the correlation between schema compliance and OCR understanding (OCRBenchV2~\citep{fu2024ocrbench} and DocVQA~\citep{mathew2021docvqa}) appears weaker with high $p$ values, the Field Match metric exhibits a moderate correlation with approx. $r=0.5$, suggesting that accurate value extraction partially benefits from text recognition capabilities. Interestingly, we do not observe strong correlations with IFEval~\citep{zhou2023instruction} and RefCOCO~\citep{yu2016modeling} benchmarks.

\begin{table}[!t]
\centering
\resizebox{0.72\linewidth}{!}{
\begin{tabular}{lccccc}
\toprule
\multirow{2}{*}{Benchmarks} &
\multicolumn{2}{c}{Schema Val.} & 
\multicolumn{2}{c}{Field Match} \\
\cmidrule(lr){2-3} \cmidrule(lr){4-5}
 & $r\uparrow$ & $p\downarrow$ & $r\uparrow$ & $p\downarrow$ \\
\midrule
IFEval~\citep{zhou2023instruction}  & -0.261 & 0.498 & -0.308 & 0.420 \\
LiveBench-250425 (IF)~\citep{white2024livebench}       & -0.257 & 0.579 & 0.539 & 0.212 \\
LiveBench-250425 (Coding)~\citep{white2024livebench}   &  0.024 & 0.960 & \colorbox{lavenderblue}{\bf 0.754} & 0.050 \\
BFCL~\citep{patilberkeley}             & \colorbox{lavender}{0.600} & 0.030 & \colorbox{lavenderblue}{\bf 0.789} & 0.001 \\
OCRBenchV2~\citep{fu2024ocrbench}      & -0.062 & 0.840 & \colorbox{lavender}{0.469} & 0.106 \\
CC\mbox{-}OCR~\citep{yang2024cc}           & -0.429 & 0.289 & -0.119 & 0.779 \\
DocVQA~\citep{mathew2021docvqa}                  & -0.163 & 0.548 & \colorbox{lavender}{0.464} & 0.070 \\
RefCOCO~\citep{yu2016modeling}         & -0.199 & 0.637 & -0.438 & 0.278 \\
MMMU (val)~\citep{yue2024mmmu}         & \colorbox{lavender}{0.555} & 0.014 & \colorbox{lavenderblue}{\bf 0.791} & 0.000 \\
MathVista~\citep{lu2023mathvista}            &  0.024 & 0.939 & \colorbox{lavender}{0.480} & 0.097 \\
MIABench~\citep{qian2024mia}           & \colorbox{lavenderblue}{\bf 0.685} & 0.090 & \colorbox{lavenderblue}{\bf 0.878} & 0.009 \\
\bottomrule
\end{tabular}
}
\caption{Pearson correlation ($r \uparrow$) and the p-value for testing non-correlation ($p \downarrow$) between structural output metrics (Schema Validation, Field Match) and external benchmarks. IFEval, LiveBench-250425 (IF) for text-only instruction following; BFCL for agentic tool use; OCRBenchV2, CC-OCR, DocVQA for text-rich image understanding; RefCOCO for image referring; MMMU for general knowledge; MathVista for math; MIABench for visual instruction following. High correlations with high confidence (low $p$ values) are highlighted in blue.} 
\label{tab:correlation_metrics}
\end{table}

\subsubsection{Schema complexity}
\begin{figure}[t!]
    \centering
    \includegraphics[width=0.76\linewidth]{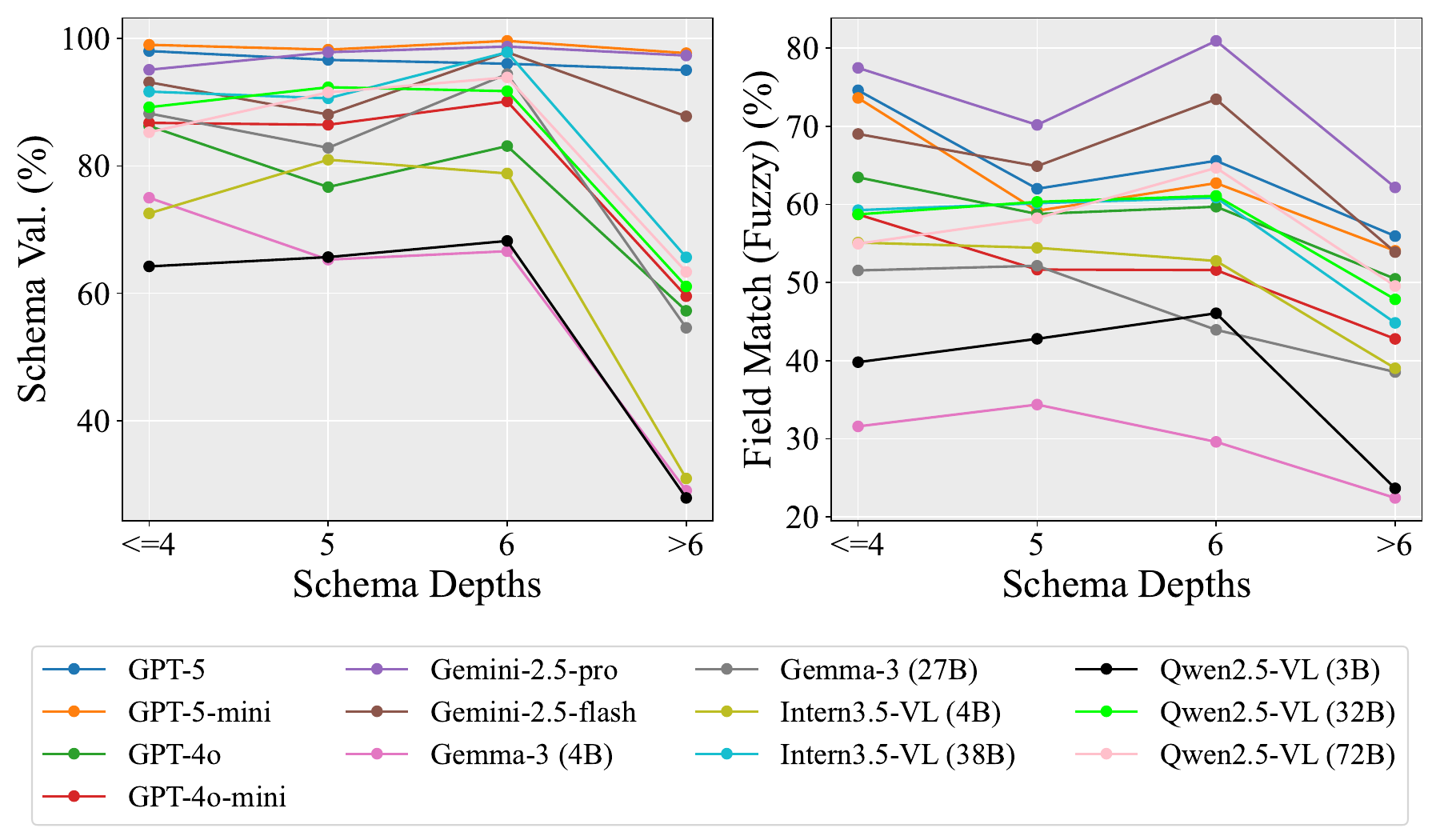}
    \caption{Model performance across different schema depths. The left panel reports schema validation accuracy, while the right shows fuzzy field-match accuracy. Number of examples per schema depth $\leq4$ with $204$ frames, $5$ with $746$ frames, $6$ with $557$ frames and $\geq6$ with $262$ frames.}
    \label{fig:ablation_schema_depth}
\end{figure}
We further analyze results by schema depth to gain a finer understanding of structured output performance. Figure \ref{fig:ablation_schema_depth} reports schema validation accuracy and fuzzy field-match accuracy across different input schema depth levels. As schema depth increases, both metrics consistently degrade for all models, suggesting that greater structural complexity poses a significant challenge to visual structured output generation. In particular, Figure~\ref{fig:ablation_schema_depth} (left) shows that GPT-5, GPT-5-mini, and Gemini-2.5-Pro sustain high schema validation accuracy above $95\%$ even at depths larger than six, whereas smaller models, such as Intern3.5-VL (4B), exhibit substantial performance drops by around $40\%$. The result highlights that small models struggle to maintain schema-compliant outputs under deeper, more nested structures, with limited complex instruction-following abilities.

\subsubsection{Inference with Structured Output API}
\begin{table}[!t]
\centering
\resizebox{0.93\linewidth}{!}{
\begin{tabular}{lcccc}
\toprule
\multirow{2}{*}{Models} & \multicolumn{2}{c}{Structured output API} & \multicolumn{2}{c}{Instruction Following} \\
\cmidrule(lr){2-3}\cmidrule(lr){4-5}
& Schema Val.& Field Match (Fuzzy) & Schema Val.& Field Match (Fuzzy) \\
\midrule
GPT-4o-mini~\citep{hurst2024gpt} & 99.75 & 51.88 & 95.19 & 56.94 \\
GPT-4o~\citep{hurst2024gpt} & 99.67 & 60.51 & 92.36 & 63.01 \\
GPT-5-mini~\citep{gpt5} & 94.52 & 60.11 & 99.92 & 61.78 \\
GPT-5~\citep{gpt5} & 95.02 & 62.67 & 99.67 & 64.52 \\
Gemini-2.5-flash~\citep{comanici2025gemini} & 95.44 & 67.99 & 98.01 & 71.41 \\
Gemini-2.5-pro~\citep{comanici2025gemini} & 96.93 & 71.83 & 98.92 & 76.17 \\
\bottomrule
\end{tabular}
}
\caption{Results from structured output APIs and instruction following prompts on a \textit{subset} of \benchname{}.}~\label{table:constrained_decoding}
\end{table}
In this section, we evaluate models using either their public structured output APIs or instruction-following prompts. To enable this comparison, we curate a subset of \benchname{} containing approximately 1.2K samples, restricted to JSON schemas supported by both OpenAI~\citep{openai_api} and Gemini~\citep{gemini_api} APIs. The results are summarized in Table~\ref{table:constrained_decoding}. Interestingly, while the GPT-4o and GPT-4o-mini models achieve slightly higher Schema Validation Accuracy when using the structured output API, their Field Match Accuracy (Fuzzy) tends to decline. In contrast, for the GPT-5 series and Gemini models, instruction-following prompts generally outperform API-based structured generation. These results suggest that although structured output APIs enforce schema compliance more reliably, they may sometimes constrain content generation, leading to less accurate field-level value predictions.

\subsection{Training Experiments}~\label{sec:training_studies}
To study how training impacts the performance of models on our \benchname{}, we further constructed a large-scale training set using the same pipeline as described in Section~\ref{sec:data_curation}, except for the human verification part. In particular, we source images from the training set of HierText~\citep{long2022towards}, AriaUI~\citep{yang2025aria} and COYO~\citep{kakaobrain2022coyo-700m}, and paired them with either our JSON schema repository or synthetic JSON schemas . We then proceed with persona-based user intent generation and response generation followed by final model-based refinement. In total, we collected 114K examples for training. We then proceed with SFT and RL experiments on this data to understand its impact. More details on training experiments are in Appendix~\ref{app:training}.

\begin{figure}[t!]
    \centering
    \includegraphics[width=0.68\linewidth]{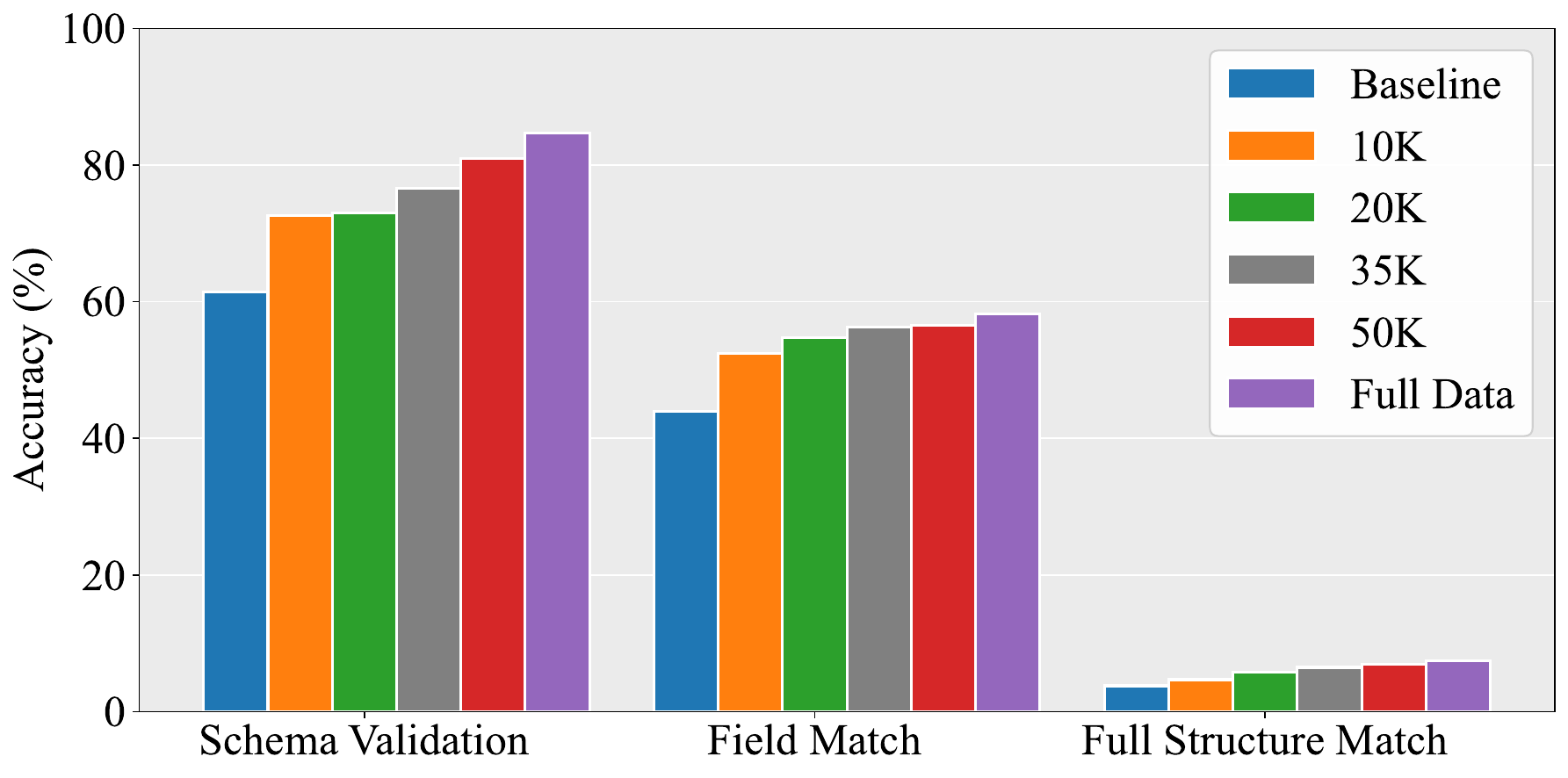}
    \caption{Performance of models trained with different scales of data. Field Match and Full Structure Match are fuzzy version.}
    \label{fig:data_scale}
\end{figure}

\subsubsection{Supervised Fine-tuning (SFT)}
We conduct our experiments on an internal 3B dense model, which is pre-trained on a mixture of image-text paired and text-only instruction following dataset. The model adopts a ViTDet-L~\citep{Li2022ExploringPV} as vision encoder and uses the AnyRes strategy~\citep{liu2024llavanext} to handle image inputs. As a starting point, we conduct supervised fine-tuning on training subsets of different sizes to understand the effect of data scaling. 
We present the baseline model and SFT models results in Figure \ref{fig:data_scale}. We see that training leads to salient gains on \benchname{}. When trained on the entire training set, our 3B model even achieves comparable performance as models that are 10x larger (in Table \ref{tab:main_results}). We also see that as we increase the data scale, the model's performance continues to improve and there is no clear sign of plateau, so we believe that larger scale of data might further improve the model performance. 

\begin{figure}[t!]
    \centering
    \includegraphics[width=0.82\linewidth]{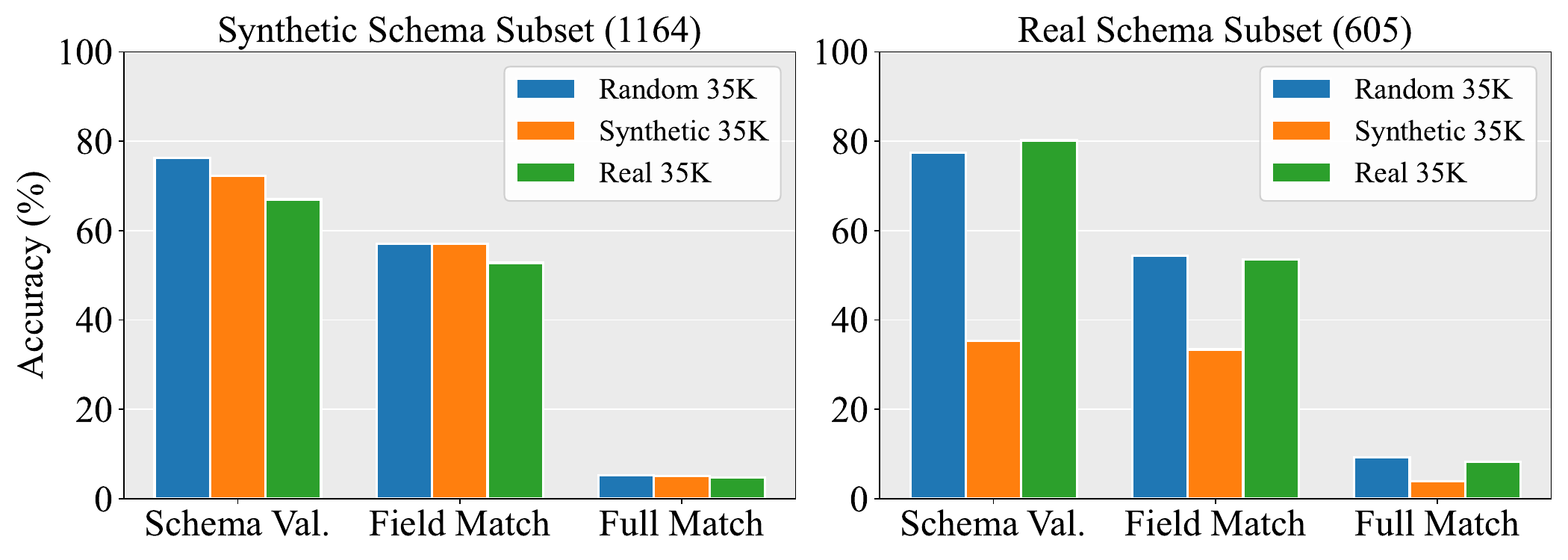}
    \caption{Performance of models on two subsets of \benchname{}. Field Match and Full Structure Match are fuzzy version.}
    \label{fig:scheme_split}
\end{figure}

\noindent\textbf{How does the model generalize to unseen schema types?} For this setting, we use the subset of training data that has synthetically generated schemas to train the model, which is about 35K examples. We compare against the models trained with either 35K examples with only real-world schemas or 35K randomly sampled data (covering both types). We present the results in Figure \ref{fig:scheme_split}. Here we further break down the results by the schema type. We see that on the subset of \benchname{} with synthetic schemas, the gaps across models are relatively smaller with the random subset achieving the best performance and real-schema subset lags behind. However, on the subset of \benchname{} with real schemas, the model trained on synthetic schema performs much worse than the other two models. Thus we think that the synthetically generated schemas still exhibit a distribution shift compared to real-world schemas, and learning from diverse set covering both types is necessary to achieve the best performance and generalization. 

\noindent\textbf{How does the model generalize to unseen image domains?} For this setting, we use only the AriaUI subset of the data to train the model, which has about 18K examples, and we compare this model with the counterpart that is trained on 20K randomly sampled data. We present the results in Figure \ref{fig:results_by_source}. Overall the model trained on AriaUI subset performances worse, and when we break down the results by image categories, we see the gap is largely explained by the Charts domain. Upon further inspection, we found that AriaUI training subset has average schema depth of 2.7, which is less than the average schema depth of the randomly sampled 20K data, which is 2.9, thus the model did not learn to handle deep schema well. As we see in Figure \ref{fig:depth_comparison}, the natural images subset has the lowest schema depth whereas charts subset has the highest schema depth, which corresponds to the results that the model trained on AriaUI subset actually performs best on natural images while lags behind the most on charts. We also find that the model often extract some correct information but fail to obey the input schema by generating a shallow or even flat structure. This behavior again suggests that training on a large and diverse set of schemas is critical to achieve strong performance. 

\subsubsection{Reinforcement Learning (RL)}

\begin{table}[t!]
\centering
\resizebox{0.66\linewidth}{!}{
\begin{tabular}{lccc}
\toprule
Models  & Schema Val. & Field Match (Fuzzy) & Full Match (Fuzzy) \\ \midrule
Baseline 3B & $58.7$ & $45.6$ & $4.4$ \\
+RLVR & $72.0$ & $47.1$ & $4.9$ \\
+SFT (14K) & $81.3$ & $54.9$ & $6.5$ \\
\hline
+SFT (50K) & $85.8$ & $56.5$ & $7.1$ \\
+SFT (50K) + RLVR & $86.6$ & $56.9$ & $6.9$ \\
\bottomrule
\end{tabular}
}
\caption{Results from baseline model and after RL/SFT training using a \textit{subset} of \benchname{} training data.} 
\label{tab:rl_results}
\end{table}

Recently, reinforcement learning with verifiable rewards (RLVR) has gained popularity within the community due to its effectiveness in improving models' performance \citep{deepseekai2025deepseekr1,yue2025does}. Verifiable rewards enjoy the benefit of clean and deterministic training signals and not needing a separate reward model. 
In our case, the correctness of structured output can also be evaluated programmatically, thus it offers a natural testbed for RLVR. In our experiments, we define the following reward function to encourage both syntactic validity and semantic correctness in generated outputs. Given a prediction $O$ and ground truth $G$, the reward $R(O, G)$ is defined as: 

\begin{equation} 
R(O, G) = \begin{cases} -0.1 & \text{if } O \text{ is invalid JSON} \\ \alpha \cdot \text{FMA}(O, G)^2 & \text{otherwise} \end{cases}, \end{equation} 

where $\text{FMA}(O, G)$ is the field match accuracy defined in Section \ref{sec:eval}, and $\alpha$ is a schema compliance multiplier: 

\begin{equation} 
\alpha = \begin{cases} 1.0 & \text{if } O \text{ is valid w.r.t the schema} \\ 0.8 & \text{otherwise} \end{cases} 
\end{equation} 
This formulation imposes a small penalty for syntactically invalid outputs, while differentiating between schema-compliant and non-compliant valid JSON through the multiplier $\alpha$. The quadratic transformation of the match score amplifies the reward signal for high-quality outputs, encouraging the policy to focus on near-correct predictions during training. We train models using Mirrow Descent Policy Optimization (MDPO) \citep{tomar2020mirror} method on a 14K subset of training data in RL experiments, and we experimented with training from a baseline checkpoint directly, as well as a checkpoint that is SFT trained on 50K non-overlapping data. To understand the performance of SFT vs RLVR, we also train a model on the same 14K data with SFT for comparison. The results are shown in Table \ref{tab:rl_results}. We see that RL training shows positive signal and it significantly boosts schema validation accuracy while also brings a small gain on field match accuracies compared to the baseline. However, compared to the SFT model counterpart the performance is still much worse. When training on top of the SFT checkpoint, we see that the gains are very small. We conjecture that baseline 3B model's capability is capped at a certain level so learning from its own sampled outputs is less effective than learning from frontier models' outputs directly. We leave further exploration on RL such as reasoning augmented training or better reward function design to future work.

\section{Conclusion}
\label{sec:conclusion}
We introduced \benchname{}, a new benchmark for evaluating the structured output capabilities of multimodal LLMs. By combining diverse visual domains with over $6.5$K JSON schemas and a unified evaluation protocol, \benchname{} provides the first systematic framework for assessing schema-grounded visual reasoning. Our experiments reveal that even the strongest frontier models still struggle with schema adherence and structural fidelity, underscoring the gap between visual understanding and reliable structured generation. Further training experiments with SFT and RLVR showcase the importance of targeted supervision for improving structured outputs.

\subsection*{Discussions and Limitations}
We identify several limitations of our work here. First, due to the high complexity of generating structured output, we rely on frontier models to automatically generate the initial benchmark labels. We applied several strategies to improve the label quality, including 1). multi-stage auto-labeling pipeline design, 2). per-stage correction from human annotators and critic models, 3). shuffling among proprietary models (e.g. GPT-5 or Gemini-2.5-Pro) at each generation stage to reduce model biases. However, there could still be some label errors which cap a model's performance. Secondly, as we discussed in the section~\ref{sec:main_results}, there could exist multiple correct ground truths for some examples and our current annotation only captures one of them. For example, semantically similar sentences for string fields, or certain flexible schema that allow different realization. Thus the models' performance might be underestimated for those cases. Finally, in our evaluation metrics design, FMA and FSMA (Equation~\ref{eq:fma} and Equation~\ref{eq:fsma}) mainly capture the recall rates of a prediction. A model which predicts many additional fields not existed in ground truth labels could still achieve high FMA or FSMA scores. We leave better metrics design as future work.

\section{Acknowledgment}
We thank Suzie Petryk, Dongxu Li, Guoli Yin, Zi-Yi Dou, Alexander Metz, and Jiarui Lu for invaluable suggestions and discussions.

\applefootnote{ \textcolor{textgray}{\sffamily Apple and the Apple logo are trademarks of Apple Inc., registered in the U.S. and other countries and regions.}}

\clearpage
\newpage
\bibliographystyle{plainnat}
\bibliography{biblio}

\clearpage
\newpage
\appendix
\clearpage
\setcounter{page}{1}

\appendix
\section{Dataset}
The original Vistext~\citep{tang2023vistext} eval dataset contains only single chart images that only have single data series, make the task less challenging. In our work, we first cluster charts into groups of size 2-6 using original chart titles bag-of-words as features and TF-IDF method to find similar charts. Then we render the chart group into one image treating each chart as a sub-plot. In total 383 charts are converted into 99 images, which are used in our data generation pipeline.

\section{Training Details}
\label{app:training}
\subsection{Training Dataset}
This section provides additional details on the training dataset used in our study (Section~\ref{sec:training_studies}). For supervised fine-tuning, we construct a training corpus sourced from the training sets of HierText~\citep{long2022towards}, AriaUI~\citep{yang2025aria}, and COYO~\citep{kakaobrain2022coyo-700m}. We balance both in-domain vs. out-of-domain data and real vs. synthetic JSON schemas. HierText serves as the in-domain dataset because its test split is used to build part of the \benchname{} evaluation set. COYO and AriaUI, in contrast, provide out-of-domain training data. For the full HierText training set (10K images), we generate one version with synthetic schemas and another with real JSON schemas. For AriaUI, we downsample to $10$K images, drawn from desktop, mobile, and web UI screens. Similar to HierText split, we generate one example with synthetic schema and another with real JSON schema for every image. Since COYO is extremely large (over 747M image–text pairs), we first categorize and then randomly downsample it to $95$K images, consisting of: $20$K infographics, $10$K handwritten images, $40$K scene-text images, $5$K maps, and $20$K tables. Among these, $80\%$ of the images are paired with real JSON schemas, while $20\%$ use synthetic schemas. Finally, we filter out noisy samples with low label quality using the critic model, ending up with $114$K training samples.

\subsection{Training hyperparameters}
In all our of SFT experiments, we train the model for around 3 epochs and we adjust the batch size (among 64, 128, 256) and training steps accordingly based on the training data size. We used learning rate of 2e-5 and max sequence length of 16k. For RL experiments, we used learning rate of 3e-7, batch size of 256. For each sample, we generate 32 rollouts and compute the advantage using the REINFORCE Leave One Out (RLOO) \citep{ahmadian2024basicsrevisitingreinforcestyle}. Again we train the model for around 3 epochs.  

\section{Error Examples}
\label{app:error_examples}
\begin{figure*}[t!]
    \centering
    \includegraphics[width=1.0\linewidth]{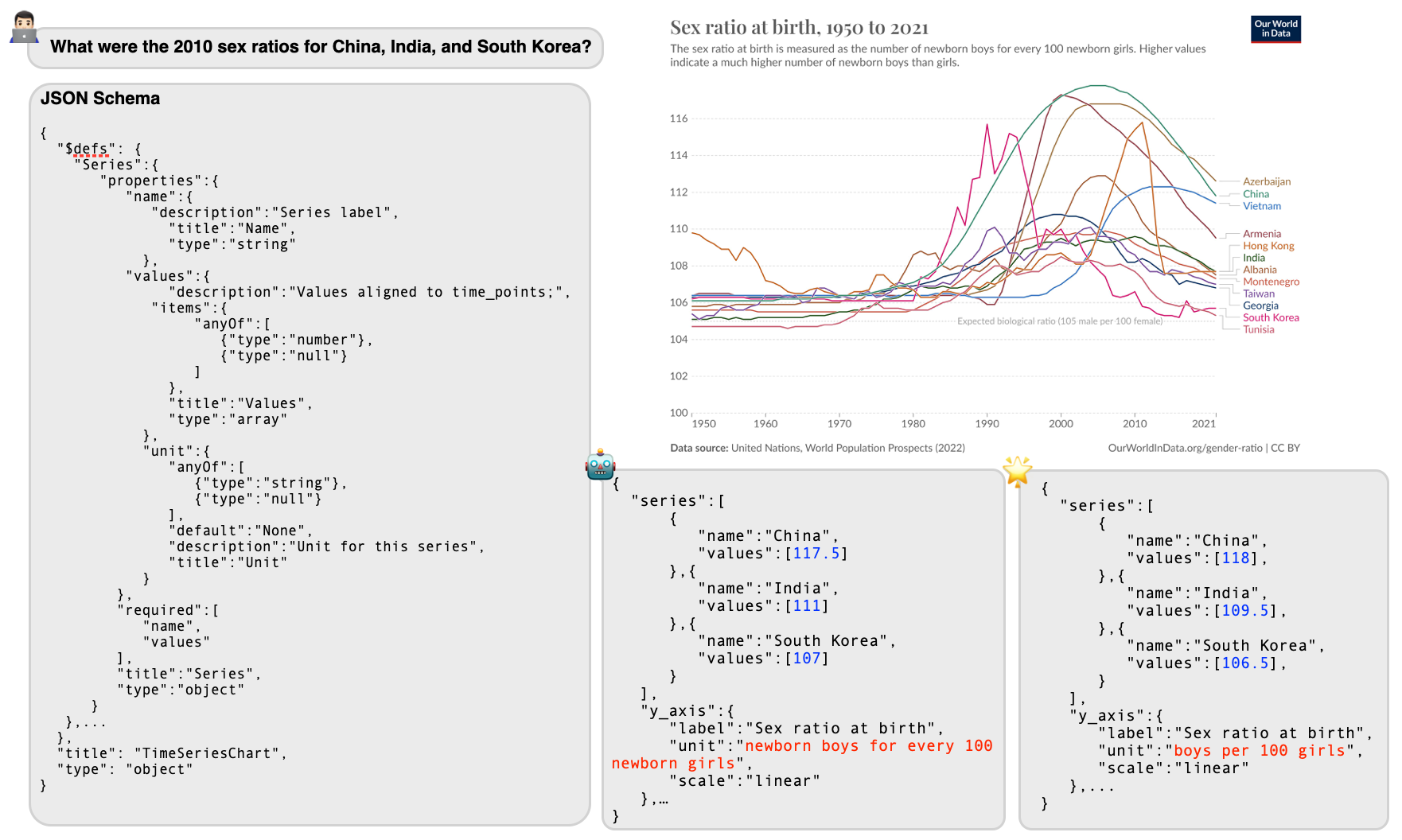}
    \caption{An example error (Chart image) from Gemini-2.5-Pro predictions. The box with Robot icon contains model predictions and the box with start icon contains ground truth. The blue font indicate that our fuzzy match metric is able to count those fields as correct, and the red font indicate mismatched field.}
    \label{fig:gemini_near_correct}
\end{figure*}

\autoref{fig:gemini_near_correct} shows an example output from Gemini-2.5-pro. We see that the model is already very good at interpreting chart data and it's predicted data values are very close to the ground truth, which are considered to be correct with fuzzy match. However, there is no explicit unit annotation on the chart, so the model has to infer the unit based on the text. Although the predicted answer also make sense, the edit distance w.r.t the ground truth already exceeds the threshold, thus it could not score full match on this example. One potential to improve our evaluation metrics is to introduce VLM-as-a-judge for fields with fuzzy match. 

\begin{figure*}[t!]
    \centering
    \includegraphics[width=1.0\linewidth]{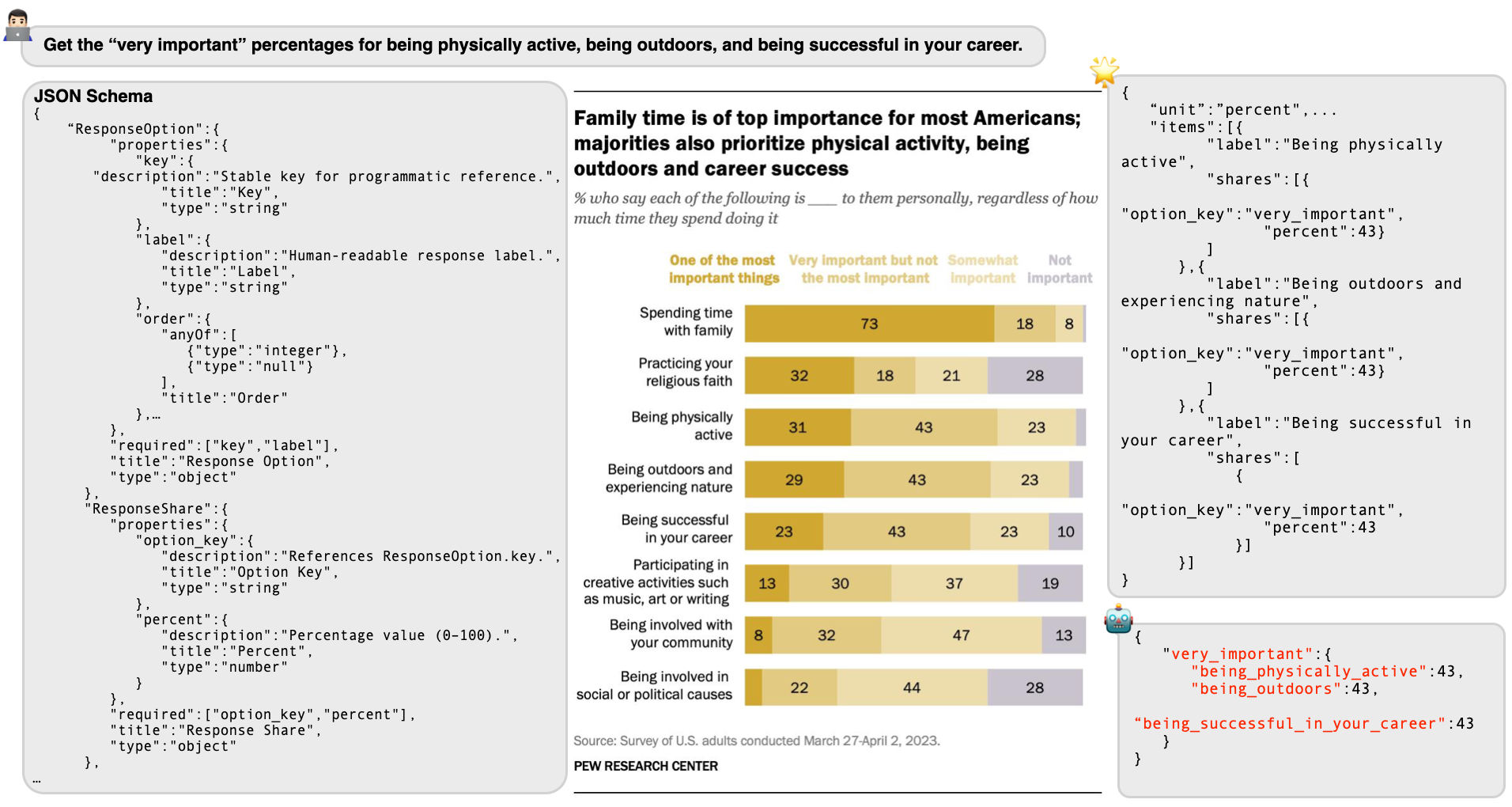}
    \caption{An example error (Chart image) from the model trained on AriaUI subset of data. Although the extracted values themselves are correct, the output does not follow the schema structure at all, leading to 0 scores on all metrics in this case.}
    \label{fig:shallow_error}
\end{figure*}

As we discussed in \autoref{sec:training_studies}, due to overall shallower schema depth in AriaUI subset of training data, the model trained on it could not learn to generalize well to deep schemas in chart domain. \autoref{fig:shallow_error} shows such an error example from the model. In this case, the extracted values are correct, but the model just hallucinated a shallow structured that is not compliant with the input schema, leading to 0 scores on all metrics. 

\begin{figure*}[t!]
    \centering
    \includegraphics[width=1.0\linewidth]{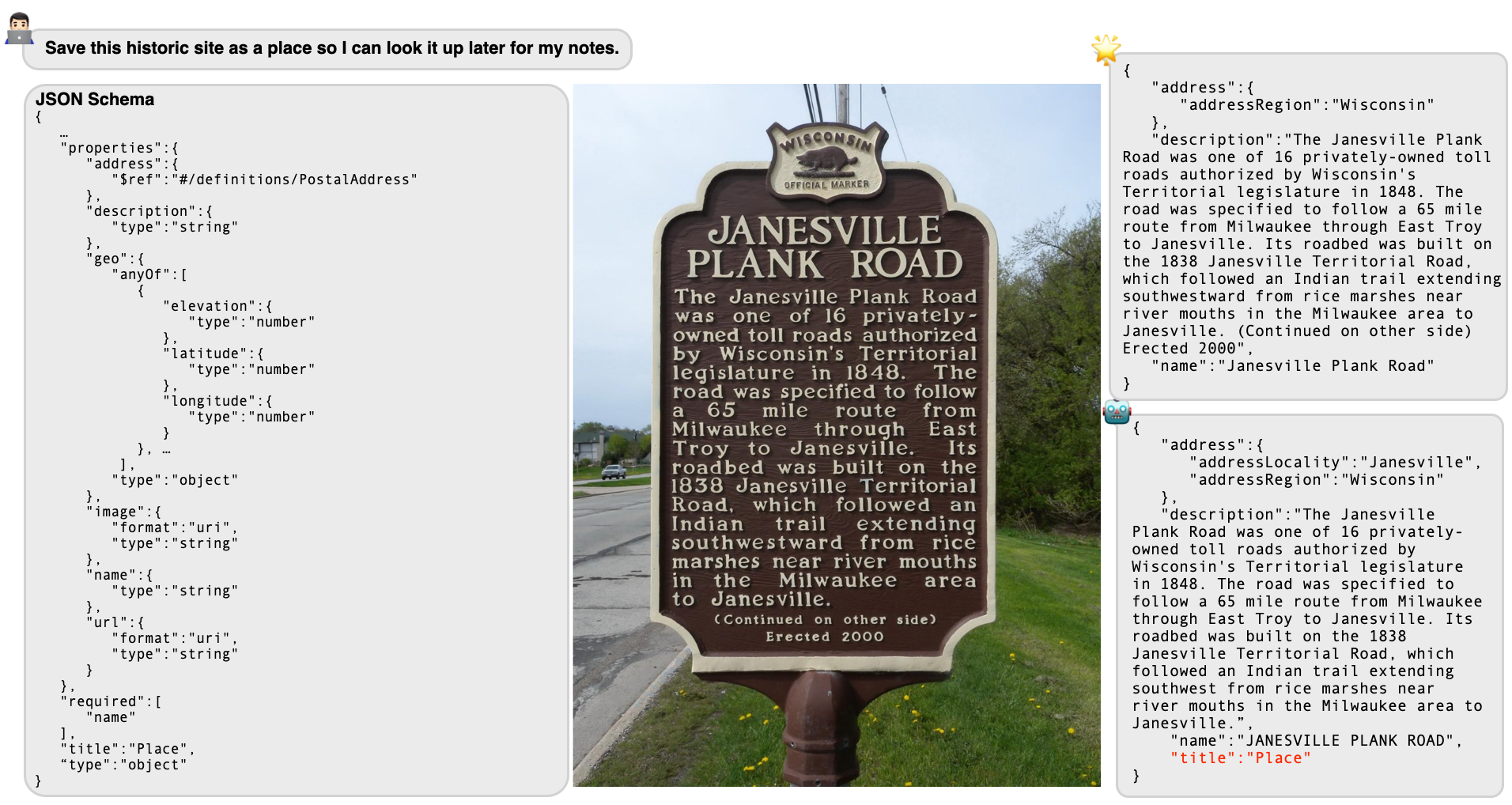}
    \caption{An example error (Natural image) from the SFT model trained on full data. This example gets partial score on field match but is invalid w.r.t. the schema}
    \label{fig:natural_error}
\end{figure*}

\begin{figure*}[t!]
    \centering
    \includegraphics[width=1.0\linewidth]{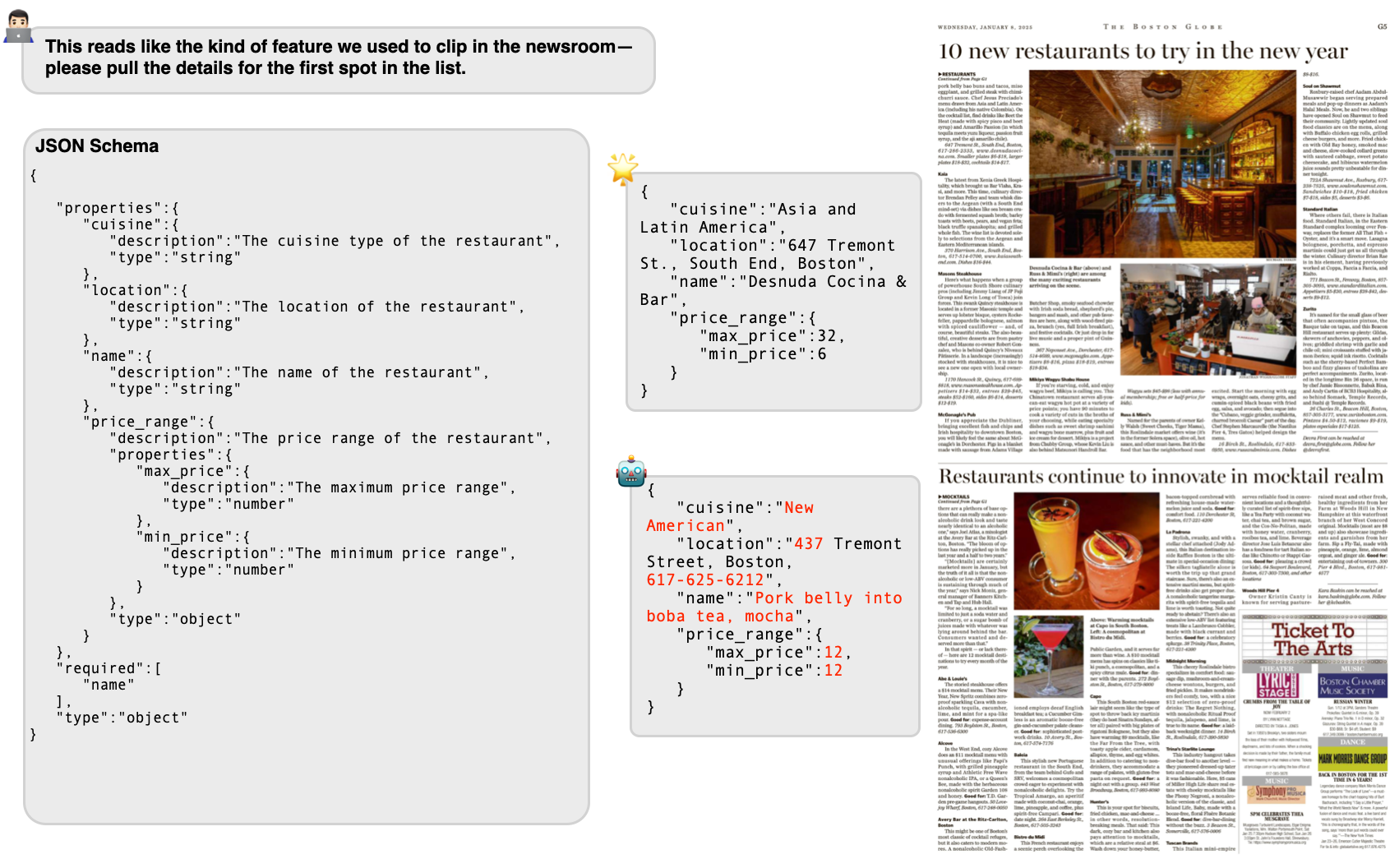}
    \caption{An example error (Document) from the SFT model trained on full data. This example gets 0 score on field match while being valid w.r.t the schema}
    \label{fig:doc_error}
\end{figure*}

\begin{figure*}[t!]
    \centering
    \includegraphics[width=1.0\linewidth]{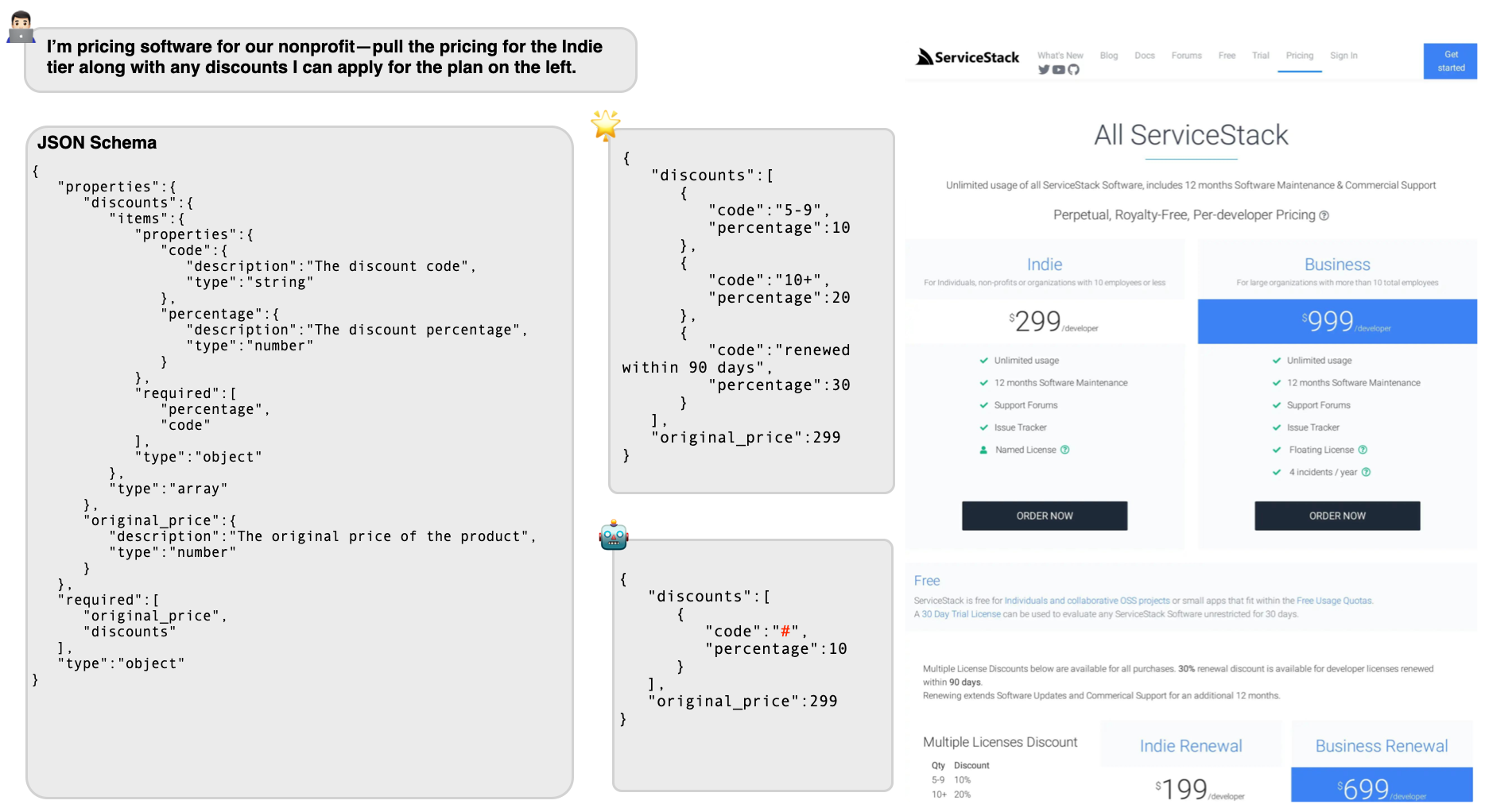}
    \caption{An example error (UI image) from the SFT model trained on full data. This example gets partial score on field match while also being valid w.r.t the schema}
    \label{fig:ui_error}
\end{figure*}

In \autoref{fig:natural_error}, \autoref{fig:doc_error}, \autoref{fig:ui_error}, we show 3 more error examples made by the SFT model trained on the entire training set. In \autoref{fig:natural_error}, we see that the model is partially correct on the generated outputs, but it mistakenly treated the title of the schema as one of the target fields, and produced an extra field which leads to 0 score on schema validation. This suggests that the model's instruction following ability could be further improved. In \autoref{fig:doc_error}, we see that the model's output is fully compliant with the schema, but all values are off compared to the ground truth. We see that the text in this image is very dense, and even for humans we have to zoom in on the first section of the page to see the details. This suggest that the visual perception ability of high-resolution images is still relatively weak. Finally, in \autoref{fig:ui_error}, we see that the model is correct in terms of schema validation and also get partial score on field match. However, it could not capture all available discounts on the page, because the information is actually scatter at different places. This hints that the general visual understanding ability that require reasoning over entire image is also a direction for improvement. 

\begin{figure*}[t!]
    \centering
    \includegraphics[width=0.96\linewidth]{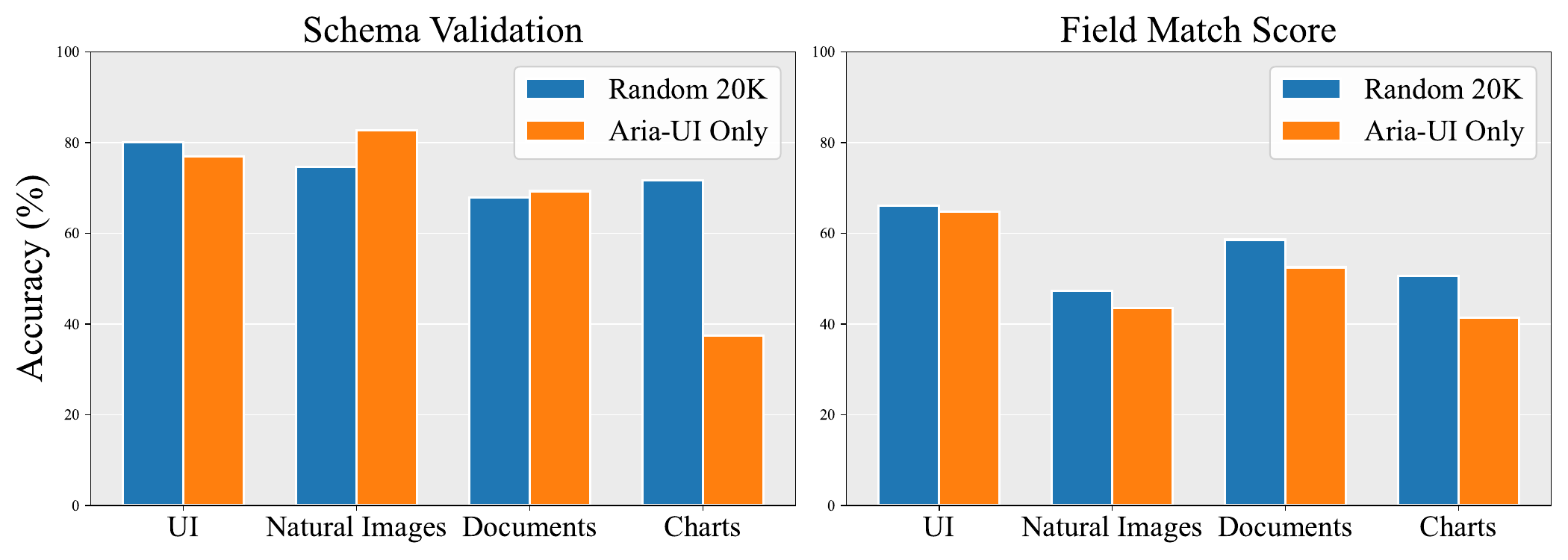}
    \caption{The results breakdown by the data categories for the model trained with 20K randomly sampled data and model trained with Aria-UI subset of the data. In the left figure, we show the schema validation accuracy and in the right figure, we show the field match accuracy (fuzzy).}
    \label{fig:results_by_source}
\end{figure*}

\section{Ablation Studies}~\label{app:ablation_studies}
\begin{figure*}[t!]
    \centering
    \includegraphics[width=0.96\linewidth]{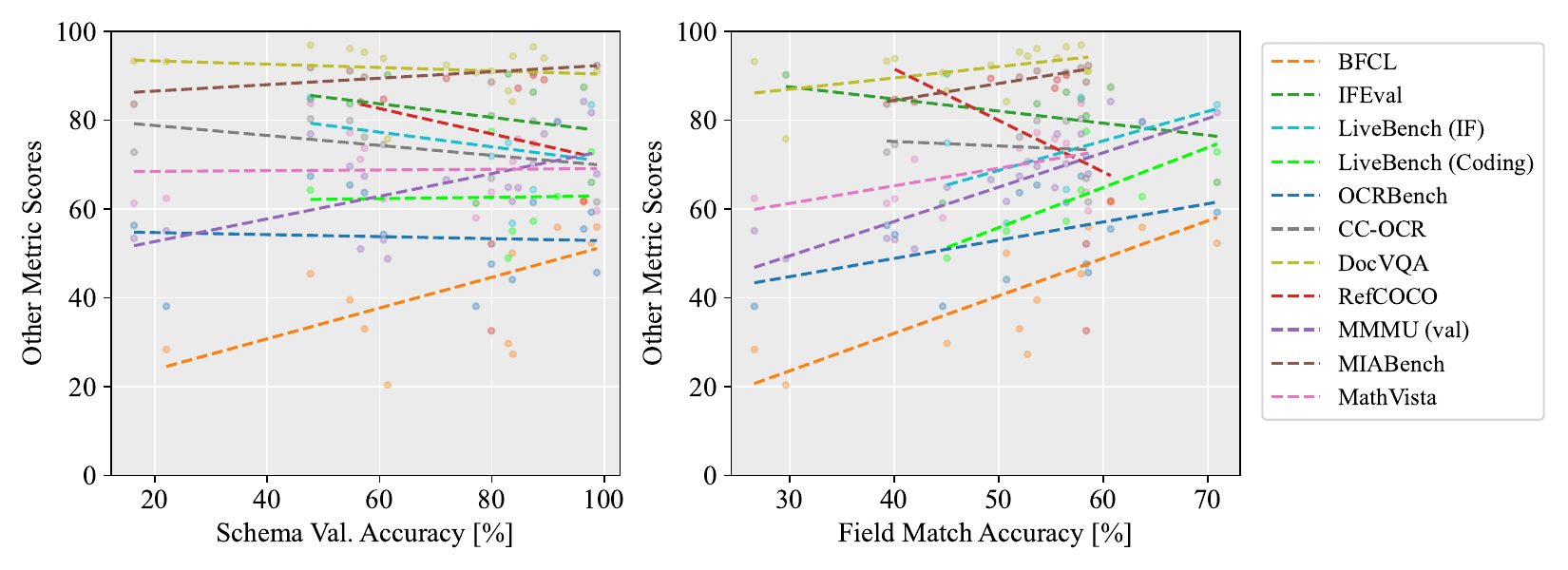}
    \caption{Metrics correlation analysis.}
    \label{fig:ablation_correlation_metrics}
\end{figure*}

Figure~\ref{fig:ablation_correlation_metrics} studies the metrics correlations between \benchname{} and other benchmarks from multiple open-sourced and proprietary models presented in Table~\ref{tab:main_results}. Their Pearson correlation coefficients are summarized in Table~\ref{tab:correlation_metrics}.

\section{Instruction Following Prompt}~\label{app:instruction_prompt}
Table~\ref{table:prompt} presents the instruction-following prompt used for structured output generation. Notably, the prompt includes explicit guidance on handling default values, particularly for fields labeled as required but not directly parsable from the input images. The same rules are consistently applied throughout both the data generation pipeline and the inference process, ensuring alignment between training and evaluation settings.

\onecolumn
\begin{center}
\begin{tcolorbox}[title=Instruction Following Prompt for Structured Output, breakable]
\ttfamily
\small
{\color{blue}\textbf{SYSTEM PROMPT}}

\textbf{Visual Structured Information Extractor}
You are an expert at extracting structured information from images according to provided schemas. Given the image and user intent, extract information that matches the provided JSON schema.

\medskip
\textbf{Constraint-Aware Default Rules (for missing/unextractable required fields)}

\textbf{General precedence}
\begin{itemize}
\item \textbf{P0 — Use explicit schema defaults}: If provided in schema, prefer these.
\item \textbf{P1 — Respect enumerations/const}: Use the declared \texttt{const} or the first \texttt{enum} value (in stable order).
\item \textbf{P2 — Apply type-specific rules while respecting schema constraints}.
\end{itemize}

\textbf{Strings}
\begin{itemize}
\item Default placeholder: \texttt{\#}.
\item Respect \texttt{minLength}/\texttt{maxLength}:
  \begin{itemize}
  \item If \texttt{minLength = m $\geq$ 1}, return exactly \texttt{m} hash characters.
  \item If \texttt{minLength = maxLength = n}, return \texttt{n} hash characters.
  \end{itemize}
\item If \texttt{pattern} forbids \#, construct the shortest valid alternative using allowed characters (\texttt{"A"}, \texttt{"0"}, etc.).
\item Only output \texttt{null} if \texttt{"null"} is explicitly allowed.
\end{itemize}

\textbf{Integers \& Numbers}
\begin{itemize}
\item Default to \texttt{0}.
\item If \texttt{minimum}/\texttt{exclusiveMinimum} is present, choose the lowest valid value.
\item If \texttt{maximum}/\texttt{exclusiveMaximum} is present and \texttt{0} violates it, choose the highest valid value under the bound.
\item If \texttt{multipleOf} is present, adjust to the nearest valid multiple.
\end{itemize}

\textbf{Booleans}
\begin{itemize}
\item Default to \texttt{false} unless otherwise constrained.
\end{itemize}

\textbf{Dates / Times}
\begin{itemize}
\item Use RFC 3339-valid placeholders:
  \begin{itemize}
  \item \texttt{"1970-01-01"} for \texttt{date}
  \item \texttt{"1970-01-01T00:00:00Z"} for \texttt{date-time}
  \item \texttt{"00:00:00Z"} for \texttt{time}
  \end{itemize}
\item If stricter patterns exist, choose the simplest valid match.
\end{itemize}

\textbf{Arrays}
\begin{itemize}
\item If \texttt{minItems = k}, return exactly \texttt{k} items, recursively filled with these same rules.
\item Ensure uniqueness if \texttt{uniqueItems:true}.
\end{itemize}

\textbf{Objects}
\begin{itemize}
\item Populate all required fields recursively.
\item Never add undeclared fields.
\end{itemize}

\medskip
\textbf{Edge-Case Notes}
\begin{itemize}
\item Never hallucinate values. Required fields must be filled with constraint-compliant placeholders.
\item Optional fields may be omitted or set to \texttt{null} if allowed.
\item If constraints are contradictory, return the closest minimally violating placeholder.
\item Stable tie-breaking: when multiple placeholder choices are possible, use the first in schema order.
\end{itemize}

\medskip
\textbf{Output Format}
\begin{itemize}
\item Return only valid JSON, wrapped in:
\begin{verbatim}
```json
{ ... }
```
\end{verbatim}
\item All required fields must appear.
\item Placeholders must respect schema constraints.
\item No comments, no extra text, no extra fields.
\end{itemize}
\medskip

{\color{blue}\textbf{USER PROMPT}}

JSON Schema:\\
\verb|```json|\\
\{schema\_json\}\\
\verb|```|

User Intent: \{user\_intent\}

Please analyze the image and extract structured information according to the schema above, honoring the user intent.\\
Return only valid JSON in a fenced \verb|```json| code block.
\end{tcolorbox}
\label{table:prompt}
\end{center}

\begin{algorithm}
\caption{Field Matching with Fuzzy Evaluation}
\label{alg:field-match}
\begin{algorithmic}[1]
\small
\Require Prediction $pred$, ground truth $gt$, labels $\mathcal{L}$
\Ensure Metrics $\mathcal{M} = \{m_{total}, m_{string}, m_{number}, m_{list}, m_{dict}\}$

\Procedure{Match}{$pred, gt, \ell$}
    \If{$\ell \neq $ ``ignore''} 
        \State $m_{total} \gets m_{total} + 1$
        
        \If{both are strings}
            \State $match \gets \begin{cases} 
                \text{EditSim}(pred, gt) > \tau_{str} & \text{if } \ell = \text{``fuzzy''} \\
                \text{Norm}(pred) = \text{Norm}(gt) & \text{otherwise}
            \end{cases}$
            \If{$match$} $m_{string} \gets m_{string} + 1$ \EndIf
        \ElsIf{both are numbers}
            \State $match \gets \begin{cases}
                |pred - gt|/\max(|gt|, \epsilon) < \tau_{num} & \text{if } \ell = \text{``fuzzy''} \\
                pred = gt & \text{otherwise}
            \end{cases}$
            \If{$match$} $m_{number} \gets m_{number} + 1$ \EndIf
        \ElsIf{both are lists of length $n$}
            \If{$\forall i \in [1,n]: $ \Call{Match}{$pred[i], gt[i], \ell[i]$}} 
                \State $m_{list} \gets m_{list} + 1$ 
            \EndIf
        \ElsIf{both are dicts}
            \State $M \gets \{k \in gt: \ell[k] \neq \text{``ignore''}\}$ \Comment{Meaningful keys}
            \If{$|M| > 0$ and $\forall k \in M: k \in pred$ and \Call{Match}{$pred[k], gt[k], \ell[k]$}}
                \State $m_{dict} \gets m_{dict} + 1$
            \EndIf
        \EndIf
    \EndIf
\EndProcedure

\State
\State \textbf{Parameters:} $\tau_{str} = 0.8$ (string similarity), $\tau_{num} = 0.05$ (relative error), $\epsilon = 10^{-9}$

\end{algorithmic}
\end{algorithm}

\end{document}